\documentclass[11pt]{article}
\usepackage{amsmath,amsfonts,graphicx,float,dsfont,color,url,caption,subcaption,booktabs,afterpage}
\usepackage{algorithm, algpseudocode}
\usepackage{epstopdf} 
\usepackage{enumerate}
\usepackage{tikz}
\usetikzlibrary{arrows, decorations.pathreplacing, shapes.geometric, shapes.arrows,scopes,svg.path,shapes.geometric,shadows}
\usetikzlibrary{calc}
\usetikzlibrary{mindmap}
\tikzset{center at zero/.style={%
execute at end picture={%
\path let \p1=(current bounding box.west),\p2=(current bounding box.east)
in ({-max(-1*\x1,\x2)},\y1) ({max(-1*\x1,\x2)},\y1);
}},
mymx/.style={ matrix of math nodes, nodes in empty cells,
    row sep=3mm, column sep=2cm,
    nodes={ minimum size=.7cm, draw, circle,
      font=\sffamily\Large\bfseries, inner sep=0.05cm } },
shbr/.style ={shade, shading=axis, left color=blue, right color=red,
    shading angle=90},
shyg/.style ={shade, shading=axis, left color=yellow, right color=green,
    shading angle=90},
shop/.style ={shade, shading=axis, left color=orange, right color=purple,
    shading angle=90},
myball/.style={shade, ball color=#1},
ashadow/.style={opacity=.25, shadow xshift=0.07, shadow yshift=-0.07},
    }

\newcommand{\reals}{\mathbb{R}}

\newcommand{\vx}{\boldsymbol x}
\newcommand{\vX}{\boldsymbol X}
\newcommand{\vw}{\boldsymbol w}

\newcommand{\norm}[2]{\left\lVert #1 \right\rVert_{#2}}

\newcommand{\ind}[1]{{\mathds{1}}_{\{#1\}}}
\newcommand{\calM}{\mathcal{M}}

\newcommand{\vzero}{\boldsymbol 0}
\newcommand{\xgboost}{XGBoost}

\DeclareMathOperator{\e}{{\rm e}}

\numberwithin{equation}{section}
\numberwithin{figure}{section}

\title{Adaptive Bayesian Reticulum}

\author{Giuseppe Nuti, Llu\'is Antoni Jim\'enez Rugama, \and Kaspar Thommen}

\begin{document}

\maketitle

\abstract{Neural Networks and Decision Trees: two popular techniques for supervised learning that are seemingly disconnected in their formulation and optimization method, have recently been combined in a single construct. The connection pivots on assembling an artificial Neural Network with nodes that allow for a gate-like function to mimic a tree split, optimized using the standard approach of recursively applying the chain rule to update its parameters. Yet two main challenges have impeded wide use of this hybrid approach: \emph{(a)} the inability of global gradient ascent techniques to optimize hierarchical parameters (as introduced by the gate function); and  \emph{(b)} the construction of the tree structure, which has relied on standard decision tree algorithms to learn the network topology or incrementally (and heuristically) searching the space at random. Here we propose a probabilistic construct that exploits the idea of a node's \emph{unexplained potential} (the total error channeled through the node) in order to decide where to expand further, mimicking the standard tree construction in a Neural Network setting, alongside a modified gradient ascent that first locally optimizes an expanded node before a global optimization. The probabilistic approach allows us to evaluate each new split as a ratio of likelihoods that balances the statistical improvement in explaining the evidence against the additional model complexity --- thus providing a natural stopping condition. The result is a novel classification and regression technique that leverages the strength of both: a tree-structure that grows naturally and is simple to interpret with the plasticity of Neural Networks that allow for soft margins and slanted boundaries.}

\section{Introduction}

Both decision trees and neural networks are popular techniques ubiquitous in the machine learning world. Neural networks tend to excel at vision and speech recognition, whilst decision trees work best in classification problems where specific subregions of the input space are associated with a different likelihood of belonging to one class (think of a heart attack risk analysis, where having a high resting heart-rate \emph{and} being older than 50 significantly increases the probability of a coronary disease).\\

Interestingly, the two techniques are complementary in that the shortcomings of one tend to be the advantages of the other. Neural networks provide for soft margin hyperplanes (if using a continuous activation function such as the classic sigmoid function) and all of the parameters are optimized at once (albeit potentially resulting in a local optimum); the main drawbacks are that (\emph{a}) the topology of the network is fixed and needs to be designed by an expert beforehand, although recent studies do cover various approaches to either pruning or growing a network (notably in \cite{cortes2016adanet} and more generally reviewed in \cite{1287896}), alongside an analysis of the theoretical bounds when introducing a regularization penalty based on the complexity of the network (in \cite{DBLP:journals/corr/CortesGKMY16}); and (\emph{b}) the network is most often a black box where the prediction is not readily understandable by inspecting the network's trained parameters. Conversely, decision trees allow for a natural method to incrementally define the structure of each tree. Arguably, the main disadvantages of decision trees are:  (\emph{a}) the inability to define a soft and slanted margin for the split (as the standard tree construction algorithms are limited to a single variable per split); and  (\emph{b}) the step-wise greedy nature of the optimization where splits closer to the root are not re-adjusted in light of newer splits (a shortcoming that a chessboard-like data set easily illustrates, see Section \ref{sec:results}).\\

The idea of a tree expressed as a neural network has been explored in various recent works: for example, in \cite{DBLP:journals/corr/abs-1711-09784} and \cite{DBLP:journals/corr/abs-1807-06699}, where Tanno \emph{et al.} provide an adaptive hybrid model that uses both gate-like functions and standard neural network components (alongside a review of similar approaches). The work by \cite{DBLP:journals/corr/abs-1807-06699} allows for the network topology to grow incrementally randomly choosing a node to extend (and if to extend it via a gate or a standard neural network transformation). Previously, the idea of using back-prorogation for decision trees via a soft margin is studied by Medina-Chico \emph{et al.} in \cite{10.1007/3-540-44795-4_30}. Biau \emph{et al.} in \cite{10.1007/s13171-018-0133-y} exploit the structure derived from a standard random forest algorithm to train a neural network-equivalent formulation (with soft margin and slanted hyperplanes), which have also been proposed in \cite{Bulo:2014:NDF:2679600.2679996} and \cite{Bertsimas2017}, where the optimization of the parameters can be either via back-propagation (in \cite{Bulo:2014:NDF:2679600.2679996}) or as a global integer problem (in \cite{Bertsimas2017}). \\

Setting the construction of a neural network in probabilistic terms has received attention of late (beyond the use of the log-likelihood as a loss function), using either Markov Chain Monte Carlo methods (as in \cite{de2003bayesian}) and Variational Bayes (in \cite{pmlr-v37-blundell15}), both more generally reviewed in \cite{mullachery2018bayesian}. Separately, Chouikhi and Alim in \cite{DBLP:journals/corr/abs-1810-13135} explore the use of the Beta function instead of the more common sigmoid activation function. Bayesian Decision Trees in \cite{Nuti19} are a fully probabilistic formulation of the data and tree-generation process and share the theoretical approach with this work.\\

As we look to join decision trees and neural networks, the obvious challenge is in how to grow the network topology to mimic the simple and effective approach in trees (which employ an exhaustive search across all possible segmentation along the input dimensions --- yet without the ability to add an angle to the separating hyperplane). The challenge in growing a neural network like a tree pivots on knowing which leaf to expand and assessing if an expansion has provided a statistically significant improvement in our model of the data --- both inextricably linked to understanding the likelihood of the data with respect to our model. As such, we introduce a key modification to the cost function of the neural network: since the tree-like formulation allows us to know the probability that an observation belongs to a particular leaf, we can simply use the posterior probability of the data (given the probability of belonging to each leaf). This has a closed form derivation for many distributions (including all the distributions in the exponential family, as detailed in \cite{Fink97acompendium}, for example). Notably, this approach removes all of the free parameters from the leaf nodes, which traditionally have had the same formulation as inner nodes (e.g. a sigmoid activation function for classification problems, etc.)\\

This new formulation allows us to tackle both of the key aspects of generating a network topology: firstly, we can prioritize expanding nodes where the data likelihood is low. Imagine a leaf node that is fully explained, i.e. where all of the points belong to the same class: there would be no point in further expanding it (and equally a leaf with very few data points is unlikely to yield a large improvement in the model). Conversely, a leaf with many heterogeneous points is a great candidate for further expansion. We call this the \emph{unexplained potential} of each leaf, which determines how likely we are to pick a leaf for expansion. Once we have expanded a node, we can examine each leaf versus its parent (by comparing the log-likelihood of the two leaves against the log-likelihood of the parent): we expect the data to be better explained by the additional complexity, with the Bayesian approach ensuring that we have sufficient statistical evidence to support the expansion.

\section{Adaptive Bayesian Reticulum} \label{sec:adp_reticulum}
Our overall objective is to estimate the probability of an outcome $y\in\reals$ given the input $\vx\in\reals^d$, namely $p(y|\vx)$. Our approach is based on a soft version of the Bayesian Decision Trees from \cite{Nuti19} which leverages the optimization approach of neural networks.

Decision trees are hierarchical models represented by directed acyclic graphs with at most one parent per node. In general, trees contain two types of nodes, internal nodes $n\in\mathcal{N}$, and terminal nodes $\ell\in\mathcal{L}$ also called leaves. The set $\mathcal{L}$ is a partition of $\reals^d$ that depends on some parameters defined in each $n\in\mathcal{N}$. We will assume all trees are binary (though one may use other types of trees). An important property of the trees is that, except for the root node $n_0$, all nodes have a unique parent. Furthermore, all internal nodes have two children connected with an edge. More specifically, decision trees are those trees whose internal nodes contain a rule activating one of the children depending on $\vx$. We name these rules the gate function $g_j$ at node $n_j$, where index $j$ is used to identify each node with its gate function. Gate functions only take values $\{0, 1\}$, defined such that the \emph{left} child is activated when $g_j(\vx) = 1$, and the \emph{right} child is activated when $1 - g_j(\vx) = 1$.  Figure \ref{fig:tree_example} provides and example of a decision tree with three internal nodes and four leaves.
\begin{figure}
\centering
\begin{tikzpicture}[x=0.8cm, y=0.8cm, square/.style={regular polygon, regular polygon sides=4}, center at zero]
    \node (x) at (0, 4.5) {$\vx$};
    \node (N_1) at (-3, 0) [circle, draw]{$n_1$};
    \node (N_0) at (0, 2.5)  [circle, draw]{$n_0$};
    \node (N_2) at (3, 0)  [circle, draw]{$n_2$};
    
    \node (L_1) at (-4.5, -3) [square, draw]{$\ell_1$};
    \node (L_2) at (-1.5, -3) [square, draw]{$\ell_2$};
    \node (L_3) at (1.5, -3)  [square, draw]{$\ell_3$};
    \node (L_4) at (4.5, -3)  [square, draw]{$\ell_4$};

    \draw[->,dashed,line width=1.5pt] (x) to [out=270, in=90] (N_0);
    \draw[->,dashed,line width=1.5pt] (x) to [out=180, in=90] (N_1);
    \draw[->,dashed,line width=1.5pt] (x) to [out=0, in=90] (N_2);
    
    \draw[->, line width=3pt, out=-160, in=60] (N_0) to node[near start, xshift=-10, yshift=0, above] {$g_0$} (N_1);
    \draw[->, line width=3pt, out=-20, in=120] (N_0) to node[near start, xshift=10, yshift=0, above] {$1-g_0$} (N_2);
    \draw[->, line width=3pt,out=210, in=90] (N_1) to node[midway,xshift=-10, yshift=0, above] {$g_1$} (L_1);
    \draw[->, line width=3pt,out=330, in=90] (N_1) to node[midway, xshift=13, yshift=0, above] {$1-g_1$} (L_2);
    \draw[->, line width=3pt,out=210,in=90] (N_2) to node[midway, xshift=-10, yshift=0, above] {$g_2$} (L_3);
    \draw[->, line width=3pt,out=330,in=90] (N_2) to node[midway, xshift=13, yshift=0, above] {$1-g_2$} (L_4);
\end{tikzpicture}
\caption{A decision tree with three internal nodes $\mathcal{N}=\{n_0, n_1, n_2\}$ and four leaves $\mathcal{L}=\{\ell_1, \ell_2, \ell_3, \ell_4\}$.} \label{fig:tree_example}
\end{figure}
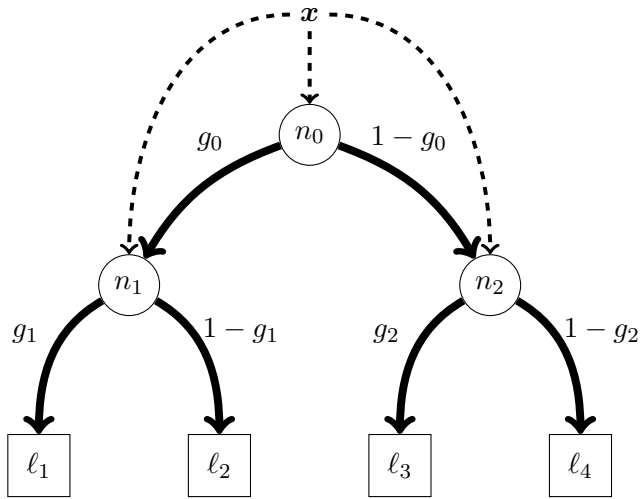

\begin{figure}
\centering
\begin{tikzpicture}[x=0.8cm, y=0.8cm, square/.style={regular polygon, regular polygon sides=4}, center at zero]
    \newcommand*{\colornodezeroleft}{red!70}
    \newcommand*{\colornodezeroright}{green!85!gray}
    
    \newcommand*{\colornodeoneleft}{blue!70}
    \newcommand*{\colornodeoneright}{orange}

    \newcommand*{\colornodetwoleft}{yellow!80!gray}
    \newcommand*{\colornodetworight}{purple!80}
    
    \newcommand*{\colorleafone}{purple!70}
    \newcommand*{\colorleaftwo}{red!45!orange!80}
    \newcommand*{\colorleafthree}{green!25!yellow!85!gray}
    \newcommand*{\colorleaffour}{green!40!purple!70}
    
    \node (x) at (0, 2) {$\vx$};
    \node (N_1) at (-3, 0) [circle, draw, left color=\colornodeoneleft, right color=\colornodeoneright]{$n_1$};
    \node (N_0) at (0, 0)  [circle, draw, left color=\colornodezeroleft, right color=\colornodezeroright]{$n_0$};
    \node (N_2) at (3, 0)  [circle, draw, left color=\colornodetwoleft, right color=\colornodetworight]{$n_2$};
    
    \node (L_1) at (-4.5, -5) [square, draw, fill=\colorleafone]{$\ell_1$};
    \node (L_2) at (-1.5, -5) [square, draw, fill=\colorleaftwo]{$\ell_2$};
    \node (L_3) at (1.5, -5)  [square, draw, fill=\colorleafthree]{$\ell_3$};
    \node (L_4) at (4.5, -5)  [square, draw, fill=\colorleaffour]{$\ell_4$};

    \draw[->, dashed, line width=1.5pt] (x) to [out=270, in=90] (N_0);
    \draw[->, dashed, line width=1.5pt] (x) to [out=180, in=70] (N_1);
    \draw[->, dashed, line width=1.5pt] (x) to [out=0, in=110] (N_2);
    
    \draw[->, \colornodezeroleft,  line width=3pt, out=-150, in=60] (N_0) to node[pos=0.1, xshift=-15] {$g_0$} (L_1);
    \draw[->, \colornodezeroleft,  line width=3pt, out=-120, in=80] (N_0) to node[pos=0.1, xshift=10] {$g_0$} (L_2);
    \draw[->, \colornodezeroright, line width=3pt, out=-60, in=100] (N_0) to node[pos=0.8, xshift=-18] {$1-g_0$} (L_3);
    \draw[->, \colornodezeroright, line width=3pt, out=-30, in=120] (N_0) to node[pos=0.82, xshift=-18] {$1-g_0$} (L_4);
    
    \draw[->, \colornodeoneleft,   line width=3pt, out=-120, in=80] (N_1) to node[pos=0.1, xshift=-8] {$g_1$} (L_1);
    \draw[->, \colornodeoneright,  line width=3pt,out=-60, in=100] (N_1) to node[pos=0.8, xshift=-20] {$1-g_1$} (L_2);
    
    \draw[->, \colornodetwoleft,   line width=3pt,out=-120,in=80] (N_2) to node[pos=0.1, xshift=-10] {$g_2$} (L_3);
    \draw[->, \colornodetworight,  line width=3pt,out=-60,in=100] (N_2) to node[pos=0.8, xshift=18] {$1-g_2$} (L_4);
\end{tikzpicture}
\caption{The neural network equivalent of the decision tree displayed in figure \ref{fig:tree_example}. Given $\vx$, each node $n_j$ emits $g_j(\vx)$ on the left and $1-g_j(\vx)$ on the right, shown as complementary colors. Leaf probabilities are the product of the received node outputs which we graphically display as mixed input colors.} \label{fig:nn_example}
\end{figure}
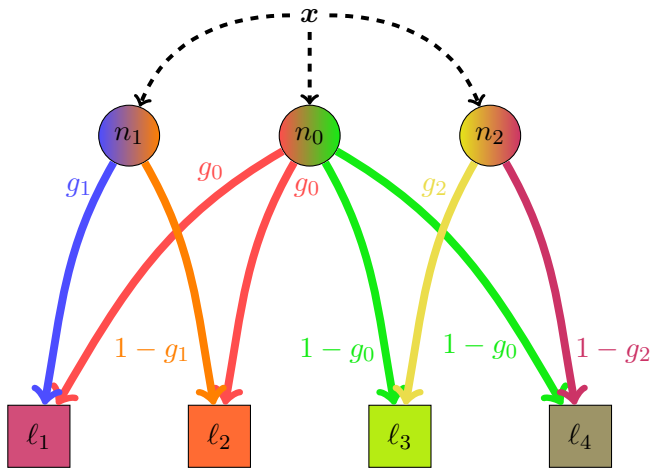

To model the probability $p(y|\vx)$ using a decision tree we define the following,
\begin{enumerate}[(a)]
\item $p(\vx\in\ell)$: probability that a known input $\vx$ belongs to leaf $\ell$, as a short hand notation for $p(\vX\in\ell|\vX=\vx)$;
\item $p(y|\vx\in\ell)$: probability of an outcome $y$ given that the known input $\vx$ belongs to leaf $\ell$, as a short hand notation for $p(y|\vX\in\ell,\vX=\vx)$.
\end{enumerate}
We can use these probabilities to rewrite $p(y|\vx)$,
\begin{equation} \label{eq:tree_probas}
p(y|\vx) = \sum_{\ell\in\mathcal{L}}p(y|\vx\in\ell)p(\vx\in\ell).
\end{equation}
Within a decision tree, the probability $p(\vx\in\ell)$ is the product of all gate functions that connect the root $n_0$ to the leaf $\ell$. For instance, in the example from Figure \ref{fig:tree_example}, $p(\vx\in\ell_1) = g_0(\vx)g_1(\vx)$, $p(\vx\in\ell_2) = g_0(\vx)(1 - g_1(\vx))$, $p(\vx\in\ell_3)=(1 - g_0(\vx))g_2(\vx)$, and $p(\vx\in\ell_4) = (1 - g_0(\vx))(1 - g_2(\vx))$. Notice that $p(\vx\in\ell)$ only takes values 0 or 1. The probability $p(y|\vx\in\ell)$ is defined per leaf and described in Section \ref{sec:bayes_reticulum}.

More generally, decision trees can be defined as a one layer neural network that includes the tree hierarchical structure constraints, \cite{10.1007/s13171-018-0133-y}. Figure \ref{fig:nn_example} shows the corresponding neural network form of the decision tree in Figure \ref{fig:tree_example}. In both formulations, the gate function for node $n_j$ is defined by a manifold $\calM$ such that $g_j(\vx)=s(d_{\calM}({\vx}))$. For this gate function, $d_{\calM}({\vx})$ is a signed distance from $\vx$ to $\calM$, and $s$, called the activation function, transforms a real value into a number between 0 and 1. In this article, we only consider $\calM$ to be from the set of affine hyperplanes, i.e. for a hyperplane with weights $\vw$, we choose $d_{\calM}({\vx}) = w_0 + \sum_{k=1}^d w_kx_k$. Decision trees are built with the Heaviside function $s(x)=\ind{x > 0}$ for activation. Crucially, a smooth $s$ allows to soften the decision trees, i.e. $p(\vx\in\ell)$ takes any value between 0 and 1 smoothly as a function of $\vw$. This allows us to optimize the parameters leveraging the back-propagation concepts from neural networks, \cite{Rum88}.

In soft decision trees, the value $p(\vx\in\ell)$ is the probability that an outcome at $\vx$ was generated by the distribution in leaf $\ell$. As a shorthand, we will keep the notation of $\vx$ belonging to $\ell$.

\subsection{Bayesian Reticulum} \label{sec:bayes_reticulum}
We can now define a Bayesian Reticulum\footnote{Inspired by the \emph{endoplasmic reticulum}, a component of organic cells. Bayesian reticula are similar in structure to endoplasmic reticula in that they both grow organically with occasional consolidation (purging) of structural elements.} for binary classification problem (regression and multiclass classification problems can naturally be extended from the definitions below). We will assume we have a set of points $\{\vx_i\}_{i=1}^m$, $\vx_i$ in $\reals^d$ with outcomes $\{y_i\}_{i=1}^m$, $y_i$ in $\{0,1\}$. The number of dichotomies into which $\vw$ can separate $\{\vx_i\}_{i=1}^m$ differently is known, \cite{Cov65}. If we assign a value to each dichotomy, an exhaustive search for an optimal dichotomy is not feasible for two reasons: the number of possible dichotomies increases exponentially with $m$, and identifying each dichotomy is hard. The problem becomes yet more complex when we consider a decision tree involving more than one hyperplane. As such, we propose a solution that consists of relaxing the activation function of decision trees to allow for the application of a gradient ascent algorithm.

A Reticulum is a decision tree whose gate functions take the form $g(\vx)=(1 + \e^{-(w_0 + \sum w_kx_k)})^{-1}$. In other words, we only consider affine hyperplanes and choose $s$ to be the sigmoid function. The Bayesian Reticulum is a Reticulum where we define the expected log-likelihood and $p(y|\vx\in\ell)$ for each leaf given the input data, as introduced in \cite{Nuti19}.

Let $|\mathcal{N}|$ be the number of internal nodes, and $|\mathcal{L}|$ the number of leaves. There are $|\mathcal{L}|^m$ different possible configurations in which each point is assigned to a unique leaf $\ell$. We can describe each configuration with an outcome $\omega\in\Omega$ that maps point $i$ to leaf $\omega(i)$. Then, each $\omega$ has a probability $p(\omega)=\prod_i p(\vx_i \in \omega(i))$ of occurring. If $\alpha$ and $\beta$ are the parameters for the Beta conjugate prior, we can obtain the posterior parameters for a leaf $\ell$ and a configuration $\omega$,
\begin{align*}
\alpha'_{\ell}(\omega) & = \alpha + \sum_{i=1}^m \ind{\omega(i)=\ell} (1 - y_i), \\
\beta'_{\ell}(\omega) & = \beta + \sum_{i=1}^m \ind{\omega(i)=\ell}y_i.
\end{align*}
The expected marginal log-likelihood for the Bayesian Reticulum is,
\begin{multline} \label{eq:exp_loglike}
\mathbb{E}_{\omega}\left[\sum_{\ell\in\mathcal{L}}\ln\left(\frac{B\left(\alpha'_{\ell}(\omega), \beta'_{\ell}(\omega)\right)}{B\left(\alpha, \beta\right)}\right) \right] \\ 
 = \sum_{\omega\in\Omega} \sum_{\ell\in\mathcal{L}}\ln\left(\frac{B\left(\alpha'_{\ell}(\omega), \beta'_{\ell}(\omega)\right)}{B\left(\alpha, \beta\right)}\right) p(\omega),
\end{multline}
where $B$ is the Beta function. Similarly, the expected probabilities at each leaf are,
\begin{equation} \label{eq:leaf_proba}
\begin{split}
p(y=0|\vx\in\ell) & = \sum_{\omega\in\Omega}\frac{\alpha'_{\ell}(\omega)}{\alpha'_{\ell}(\omega) + \beta'_{\ell}(\omega)} p(\omega), \\
p(y=1|\vx\in\ell) & = \sum_{\omega\in\Omega}\frac{\beta'_{\ell}(\omega)}{\alpha'_{\ell}(\omega) + \beta'_{\ell}(\omega)} p(\omega).
\end{split}
\end{equation}
Note that the probabilities in equation \eqref{eq:leaf_proba} are not free parameters and are fully determined by $p(\vx_i\in\ell)$ and $y_i$ (unlike standard neural networks where the last layer is parametrized as the \emph{de facto} probability of belonging to a class and optimized by gradient ascent).

Our goal is to find the optimal Bayesian Reticulum given the input data, i.e. to maximize equation \eqref{eq:exp_loglike}. The summation over $|\mathcal{L}|^m$ is not computationally feasible: as such, we optimize a lower bound of \eqref{eq:exp_loglike}. Because the log-beta function is convex, \cite{Dra00}, we apply the Jensen's inequality \cite{Jen06} and obtain,
\begin{equation}  \label{eq:approx_loglike}
c = \sum_{\ell\in\mathcal{L}}\ln\left(\frac{B\left(\alpha'_{\ell}, \beta'_{\ell} \right)}{B\left(\alpha, \beta\right)}\right),
\end{equation}
with
\begin{align*}
\alpha'_{\ell} & = \mathbb{E}_{\omega}\left[\alpha'_{\ell}(\omega)\right] = \alpha + \sum_{i=1}^{m} p(\vx_i\in\ell) (1 - y_i), \\
\beta'_{\ell} & = \mathbb{E}_{\omega}\left[\beta'_{\ell}(\omega)\right] = \beta + \sum_{i=1}^{m} p(\vx_i\in\ell) y_i.
\end{align*}
The probabilities at each leaf can also be approximated,
\begin{equation} \label{eq:leaf_approx_proba}
\begin{split}
p(y=0|\vx\in\ell) & = \frac{\alpha'_{\ell}}{\alpha'_{\ell} + \beta'_{\ell}}, \\
p(y=1|\vx\in\ell) & = \frac{\beta'_{\ell}}{\alpha'_{\ell} + \beta'_{\ell}}.
\end{split}
\end{equation}

Finally, we define the value $c_\ell=\ln(B(\alpha'_\ell, \beta'_\ell) / B(\alpha, \beta))$ as the leaf \emph{unexplained potential} --- i.e. the lower bound on the (log) probability of the data at a particular leaf (which will be useful as we look to explore which leaf to expand further). The intuition behind this definition is that the smaller the value of $c_\ell$, the more likely we are to find a new set of weights $\vw$ at leaf $\ell$ that improves the Reticulum expected log-likelihood.

\subsection{Adaptive Construction of Bayesian Reticula} \label{sec:construction}

The main objective in constructing a Bayesian Reticulum is to maximize the bound $c$ from equation \eqref{eq:approx_loglike}. We do not assume any \emph{a priori} knowledge of the structure of the Reticulum. Instead, we discover the structure adaptively based on the input data. During this construction, we will attempt to extend randomly selected leaves by sampling and optimizing new node weights for these leaves.

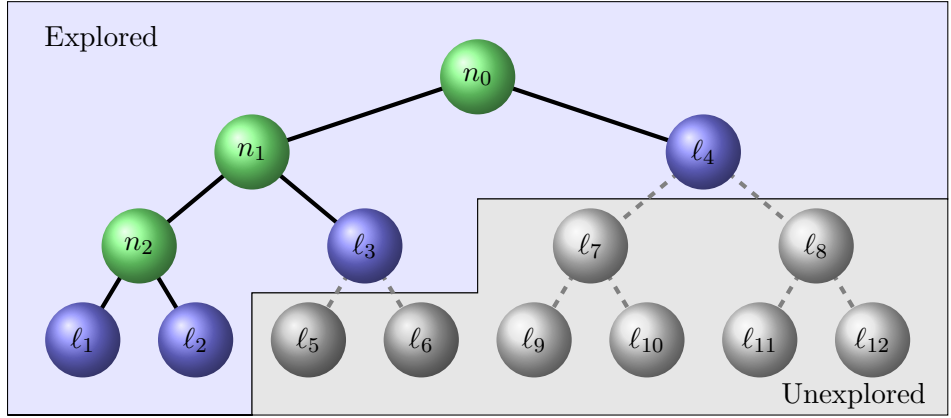
\begin{figure}
\centering
\begin{tikzpicture}
    \coordinate (n0) at (0,0) ;
    \coordinate (n1) at (-3, -1) ;
    \coordinate (n3) at (-4.5, -2.25) ;
    \coordinate (l7) at (-5.25, -3.5) ;
	\coordinate (l11) at (-3.75, -3.5) ;
    \coordinate (l5) at (-1.5, -2.25) ;
	\coordinate (l9) at (-2.25, -3.5) ;
	\coordinate (l13) at (-0.75, -3.5) ;
    \coordinate (l1) at (3, -1) ;
	\coordinate (l4) at (1.5, -2.25) ;
    \coordinate (l8) at (0.75, -3.5) ;
    \coordinate (l12) at (2.25, -3.5) ;
    \coordinate (l6) at (4.5, -2.25) ;
    \coordinate (l10) at (3.75, -3.5) ;
	\coordinate (l14) at (5.25, -3.5) ;
	
	\draw[fill=blue!10] (-6.25,1) -- (-6.25,-4.5) -- (-3,-4.5) -- (-3,-2.875) -- (0,-2.875) -- (0,-1.625) -- (6.25,-1.625) -- (6.25, 1) -- cycle ;
	\node at (-5.0,0.5) {Explored} ;
	\draw[fill=gray!20] (-6.25,-4.5) -- (-3,-4.5) -- (-3,-2.875) -- (0,-2.875) -- (0,-1.625) -- (6.25,-1.625) -- (6.25, -4.5) -- cycle ;
	\node at (5.0, -4.25) {Unexplored} ;

	\draw[ultra thick] (n0) -- (n1);
	\draw[ultra thick] (n0) -- (l1);
	\draw[ultra thick] (n1) -- (n3);
	\draw[ultra thick] (n1) -- (l5);
	\draw[ultra thick] (n3) -- (l7);
	\draw[ultra thick] (n3) -- (l11);
	\draw[ultra thick, dashed, gray] (l5) -- (l9);
	\draw[ultra thick, dashed, gray] (l5) -- (l13);
	\draw[ultra thick, dashed, gray] (l1) -- (l4);
	\draw[ultra thick, dashed, gray] (l1) -- (l6);
	\draw[ultra thick, dashed, gray] (l4) -- (l8);
	\draw[ultra thick, dashed, gray] (l4) -- (l12);
	\draw[ultra thick, dashed, gray] (l6) -- (l10);
	\draw[ultra thick, dashed, gray] (l6) -- (l14);

	\shade[myball=green!50]  (n0) circle (.5cm) node {$n_0$};
	\shade[myball=green!50]  (n1) circle (.5cm) node {$n_1$};
	\shade[myball=green!50]  (n3) circle (.5cm) node {$n_2$};
	\shade[myball=blue!50]  (l7) circle (.5cm) node {$\ell_1$};
	\shade[myball=blue!50]  (l11) circle (.5cm) node {$\ell_{2}$};
	\shade[myball=blue!50]  (l5) circle (.5cm) node {$\ell_3$};
	\shade[myball=gray!50]  (l9) circle (.5cm) node {$\ell_5$};
	\shade[myball=gray!50]  (l13) circle (.5cm) node {$\ell_{6}$};

	\shade[myball=blue!50]  (l1) circle (.5cm) node {$\ell_4$};
	\shade[myball=gray!25]  (l4) circle (.5cm) node {$\ell_7$};
	\shade[myball=gray!25]  (l8) circle (.5cm) node {$\ell_9$};
	\shade[myball=gray!25]  (l12) circle (.5cm) node {$\ell_{10}$};
	\shade[myball=gray!25]  (l6) circle (.5cm) node {$\ell_8$};
	\shade[myball=gray!25]  (l10) circle (.5cm) node {$\ell_{11}$};
	\shade[myball=gray!25]  (l14) circle (.5cm) node {$\ell_{12}$};
\end{tikzpicture}
\caption{Reticulum with a maximum of four levels. At this stage, we extended nodes $n_0, n_1$, and $n_2$. The current leaves are $\ell_1, \ell_2, \ell_3$, and $\ell_{4}$. Because there is a maximum of four levels, only $\ell_3$ and $\ell_4$ can be extended. The gray region of the Reticulum has not been explored yet.} \label{fig:ret_const}
\end{figure}

We provide the general algorithm to construct Bayesian Reticula in Algorithm \ref{alg:reticulum_algo}. First, we initialize the Reticulum starting at the root node (leaf). As the number of leaves grows, we choose the leaf we want to extend randomly proportional to the unexplained potential, i.e. according to the probabilities
\begin{equation} \label{eq:leaf_prob}
\frac{c_\ell}{\sum_{\ell^*\in\mathcal{L}}c_{\ell^*}}, \quad \text{for leaf } \ell.
\end{equation}
For the chosen leaf, we sample the weights uniformly over the unit sphere and locally optimize the weights, keeping the rest of the network parameters constant. After the local optimization, we perform a global optimization\footnote{The log-likelihood function bound of the Bayesian Reticulum defined in equation \eqref{eq:approx_loglike} is differentiable with respect to the node weights. Thus, one can apply any gradient ascent algorithm to optimize the weights. As proposed for neural networks, \cite{Rum88}, we provide the details of a back-propagation algorithm in Appendix \ref{app:backprop}.} and finally, assess every leaf to determine whether it should be pruned. In other words, we keep those leaves that improve the marginal log-likelihood with respect to their parent by a minimum amount. Figure \ref{fig:ret_const} shows an example of a partially constructed Reticulum.

\begin{algorithm}
\caption{Construction of Bayesian Reticula.}
\label{alg:reticulum_algo}
\begin{algorithmic}[1]
\Procedure{construct}{}
    \State $a \gets 0$, initialize the Reticulum \label{algo:line:init}
    \While {$a < \text{max\_attempts}$}
        \State $j \gets \text{choose the index of node to extend}$ \label{algo:line:choose}
        \State Sample the weights for node $j$ \label{algo:line:sample_weights}
        \State Train node $j$'s weights using gradient ascent\label{algo:line:train1}
        \State Train the full tree's weights using gradient ascent\label{algo:line:train2}
        \State Prune the Reticulum \label{algo:line:prune}
        \State $a \gets a + 1$
    \EndWhile
\EndProcedure
\end{algorithmic}
\end{algorithm}

The construction of Bayesian Reticula is sensitive to steps \ref{algo:line:choose} to \ref{algo:line:prune}. In particular, the gradient ascent convergence properties will affect the effectiveness of Algorithm \ref{alg:reticulum_algo}. These properties are easier to explain in polar coordinates, see Appendix \ref{app:polar}. As such, we define $r$ as the \textit{stiffness} of a node, which is soft when $r$ is small, and stiff when $r$ is large.

In practice, if a node is stiff the sigmoid $s$ has been scaled to look like the Heaviside activation function. In this case, the value $s$ for a point $i$ and node $j$, namely $s_j^i$, becomes 0 or 1 and the gradient is close to $\vzero$, see equation \eqref{eq:partial_a_w}. When $r$ is small, the sigmoid becomes flatter, centered around $1/2$. For soft nodes, the values of $s_j^i$ are close to 1/2 and the gradient values for all coordinates but $r$ are close to 0. In this case, the intuition is that all points have equal weights and the most sensitive variable to the Bayesian Reticulum log-likelihood bound $c$ is the stiffness.

Furthermore, the hierarchical structure of the Reticulum causes lower level nodes to have a smaller impact on $c$ as opposed to upper level nodes. For instance, the value of $g_0$ affects $p(\vx\in\ell)$ for all leaves while $g_1$ only affects the leaves on one side of the root node $n_0$. If we perform a global gradient ascent optimization to the Reticulum, we observe three important convergence properties: 1) the lower level nodes will barely move with respect to the upper level nodes, 2) stiff nodes will not move, and 3) soft nodes need to stiffen before finding a hyperplane that improves the unexplained potential $c_\ell$. Considering these properties, the training step is performed in two phases. First, we perform a local optimization by only updating the current extended node weights (line \ref{algo:line:train1}). Second, we finish with a global optimization that accounts for all nodes (line \ref{algo:line:train2}). The first phase is the optimization that regular decision trees perform, while the second is what neural networks perform. This is what places the Reticula as a mixture of both techniques.

After each node extension, we verify whether all nodes are still relevant. Starting from the parents whose children are both leaves, we verify that the sum of the children's unexplained potentials is better than their parent unexplained potential, i.e. without the split. If the siblings' value did not improve the parent value by a minimum factor, we prune the parent by setting it to a leaf and repeat the process. In some cases, during the global optimization, points can move from one leaf to another leaving some empty leaves. These empty leaves are an example where a parent would be pruned because the split does not improve the parent's unexplained potential.

We provide a detailed view for the implementation of Algorithm \ref{alg:reticulum_algo} in Appendix \ref{app:implementation}.

\section{Results} \label{sec:results}
We compare our Adaptive Bayesian Reticulum to a decision tree, a neural network, and an \xgboost{} classifier, \cite{xgboost16}.  We apply these techniques to three examples we can visually explain: the recovery of a soft sphere $\mathbb{S}^1$, the recovery of a cross, and the Ripley data set \cite{Rip94a}. We chose these examples because despite their simplicity, some of the models obtained will still suffer from complex and unexplainable topologies. The Adaptive Bayesian Reticulum implementation and figure scripts are provided in the GitHub repository \cite{ABR19}.

The hyperparameters of all techniques have been optimized using the Optuna hyperparameter tuning library \cite{optuna_2019}. We chose to minimize the log-loss (or cross-entropy) metric using a 5-fold cross-validation to avoid overfitting and we report the average out-of-sample log-loss across the folds. The log-loss for a binary outcome $y$ is defined as $-\left( y\log(p) + (1-y)\log(1-p) \right)$ where $p$ is the model's predicted probability of $y$ belonging to class $1$.

\subsection{Recovery of a Sphere}
In this example we train a Reticulum with $5\,000$ points generated under $\vX=((1 + 0.3N)\cos(2\pi U), (1 + 0.3N)\sin(2\pi U))^T$, where $U\sim \text{Uniform}(0,1)$ and $N\sim\text{Normal}(0,1)$ are independent. The outcome $Y$ follows a Bernoulli distribution with probability $p$. This probability is a function of the Euclidean distance between $\vX$ and $\mathbb{S}^1$, namely $p=1/(1+\e^{-10\norm{\vX}{2}})$.

\begin{figure}
\centering
\includegraphics[width=0.45\textwidth]{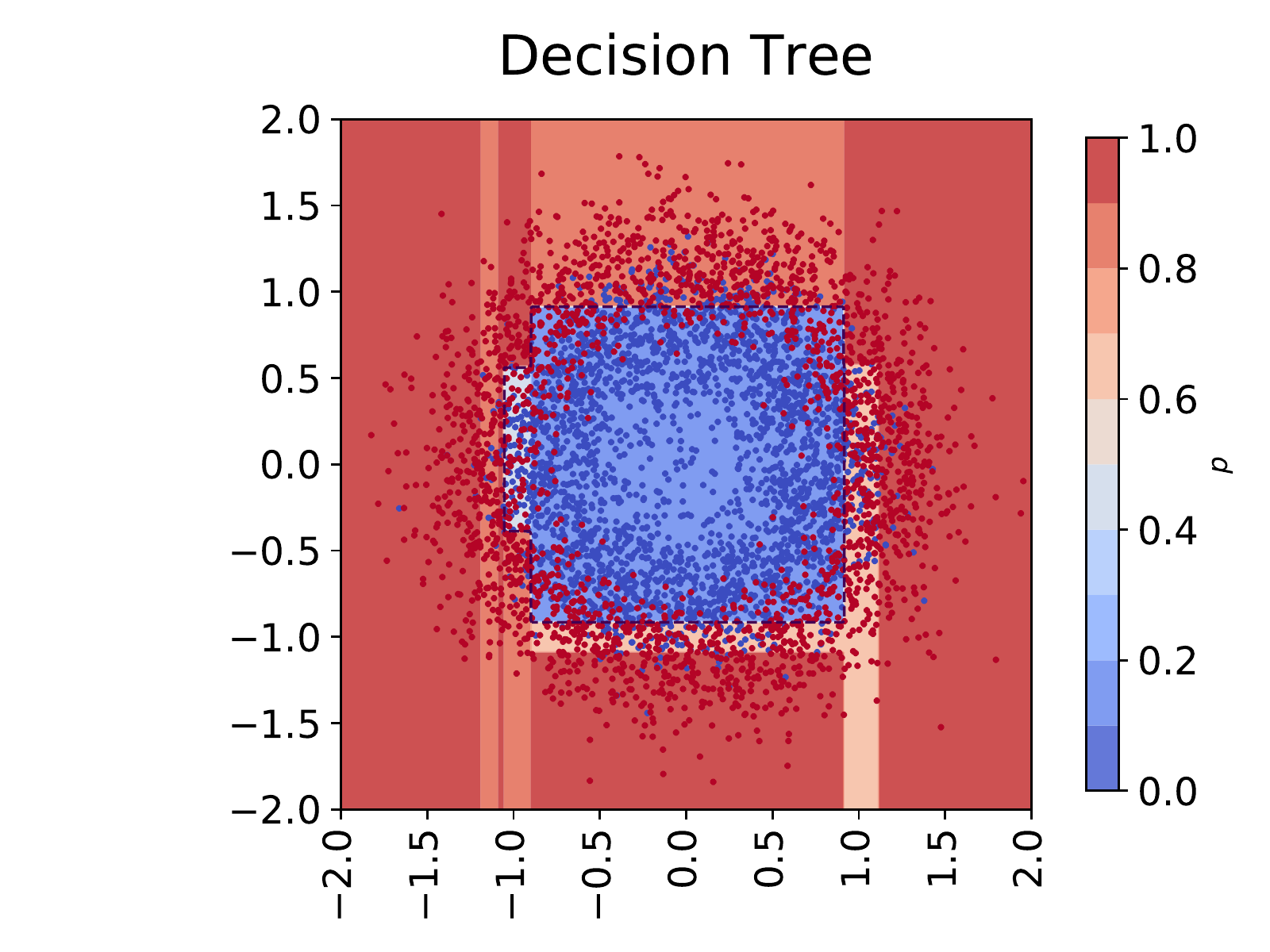}
\includegraphics[width=0.45\textwidth]{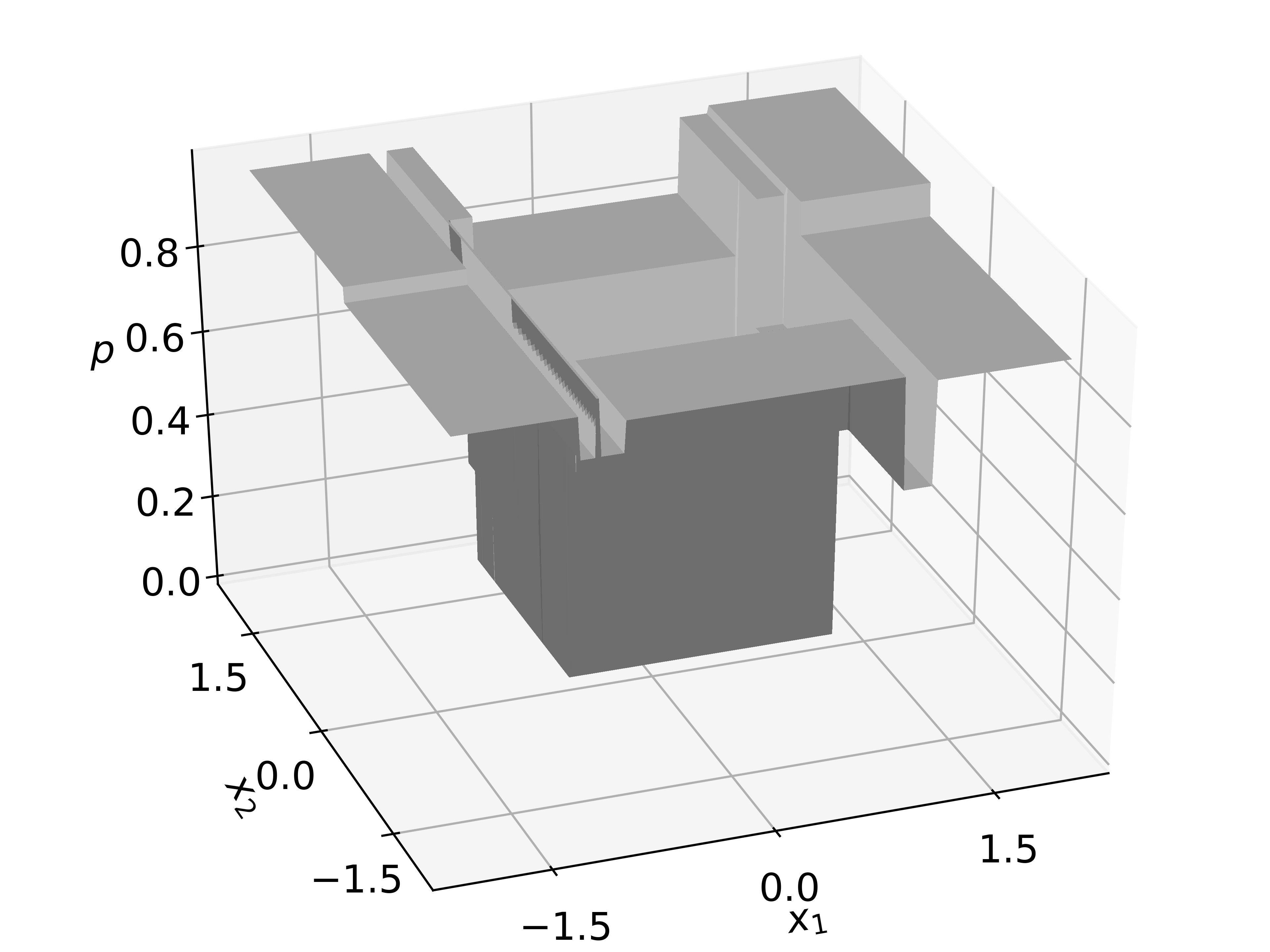} \\
\includegraphics[width=0.45\textwidth]{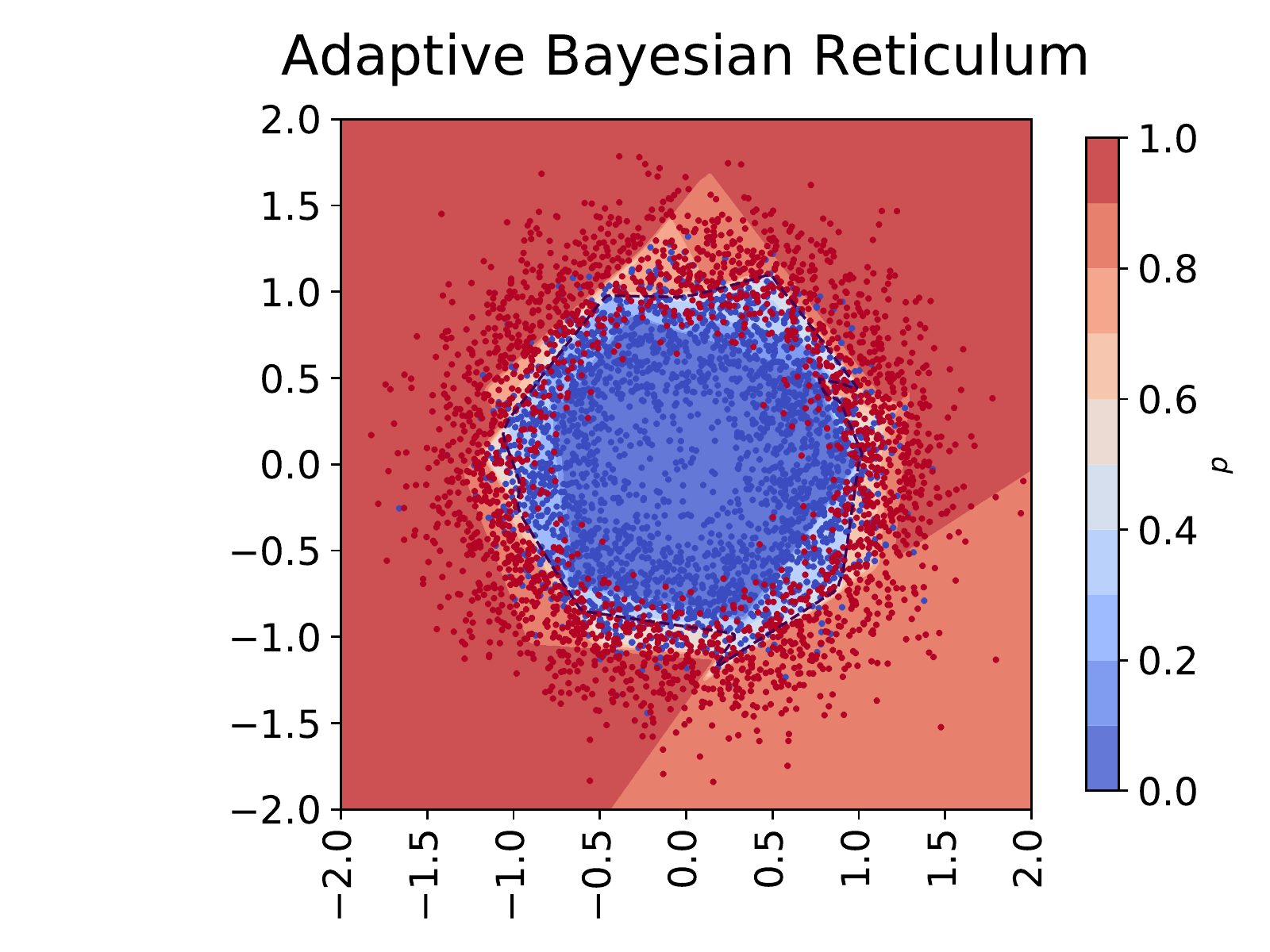}
\includegraphics[width=0.45\textwidth]{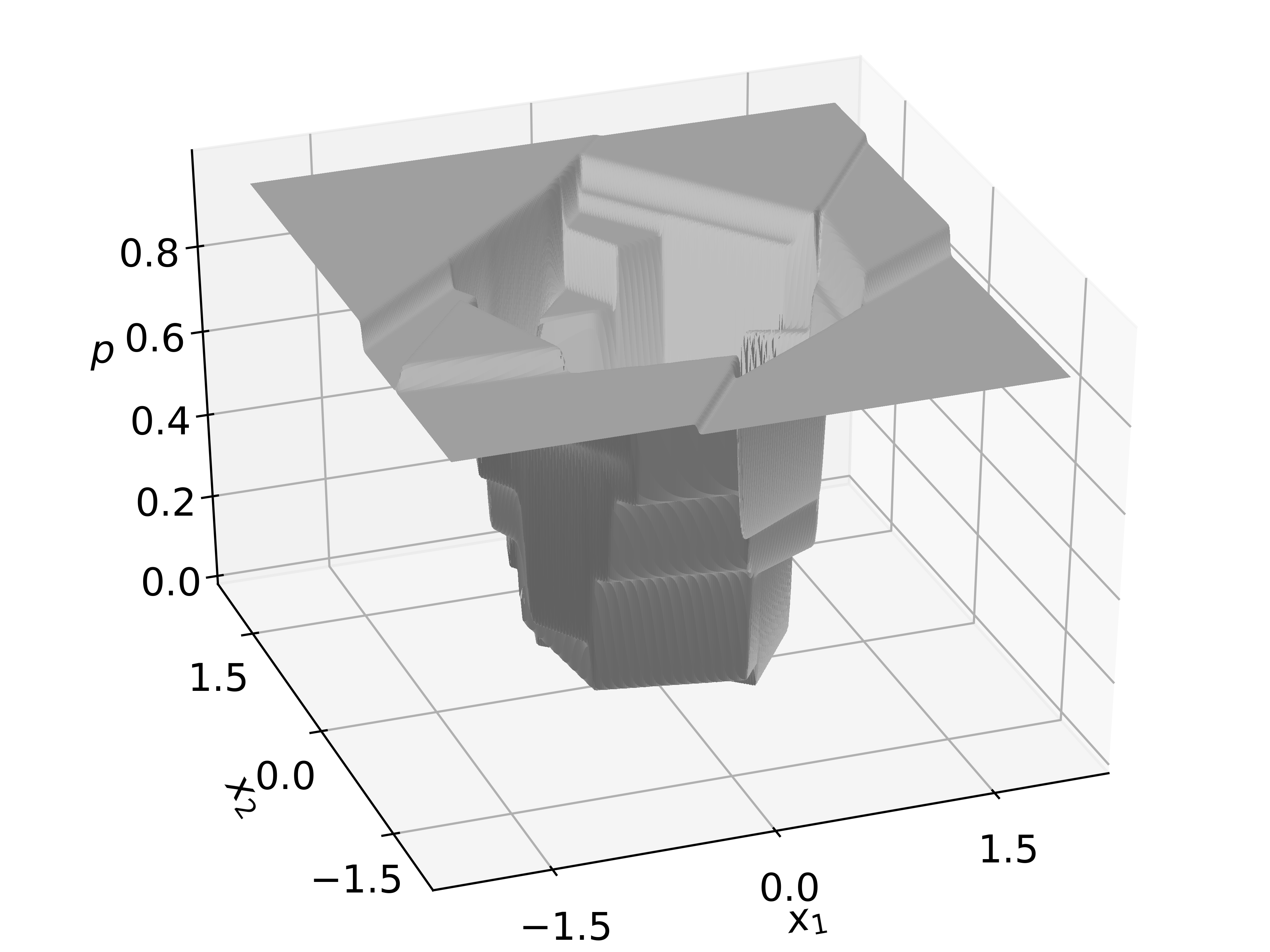} \\
\includegraphics[width=0.45\textwidth]{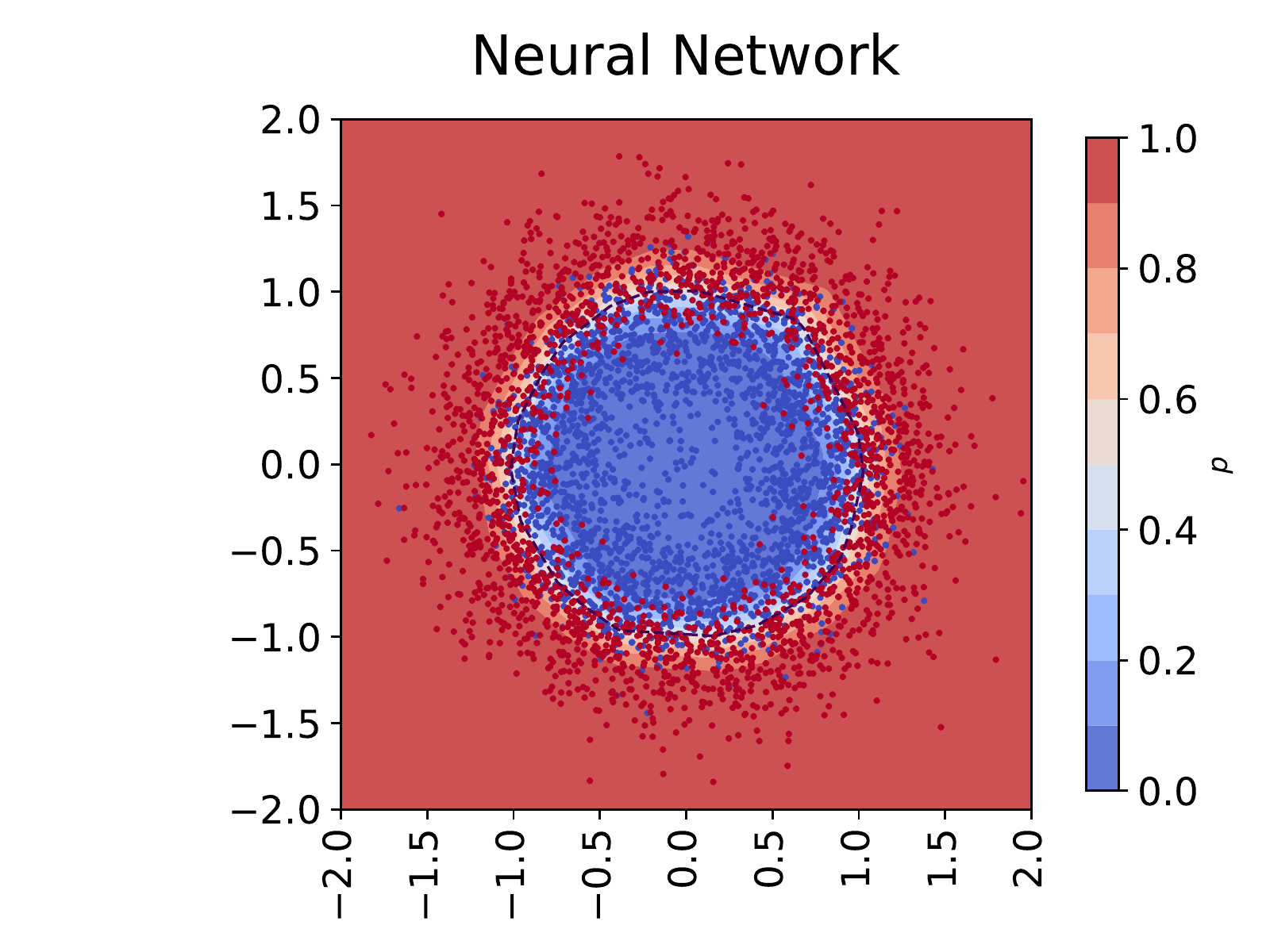}
\includegraphics[width=0.45\textwidth]{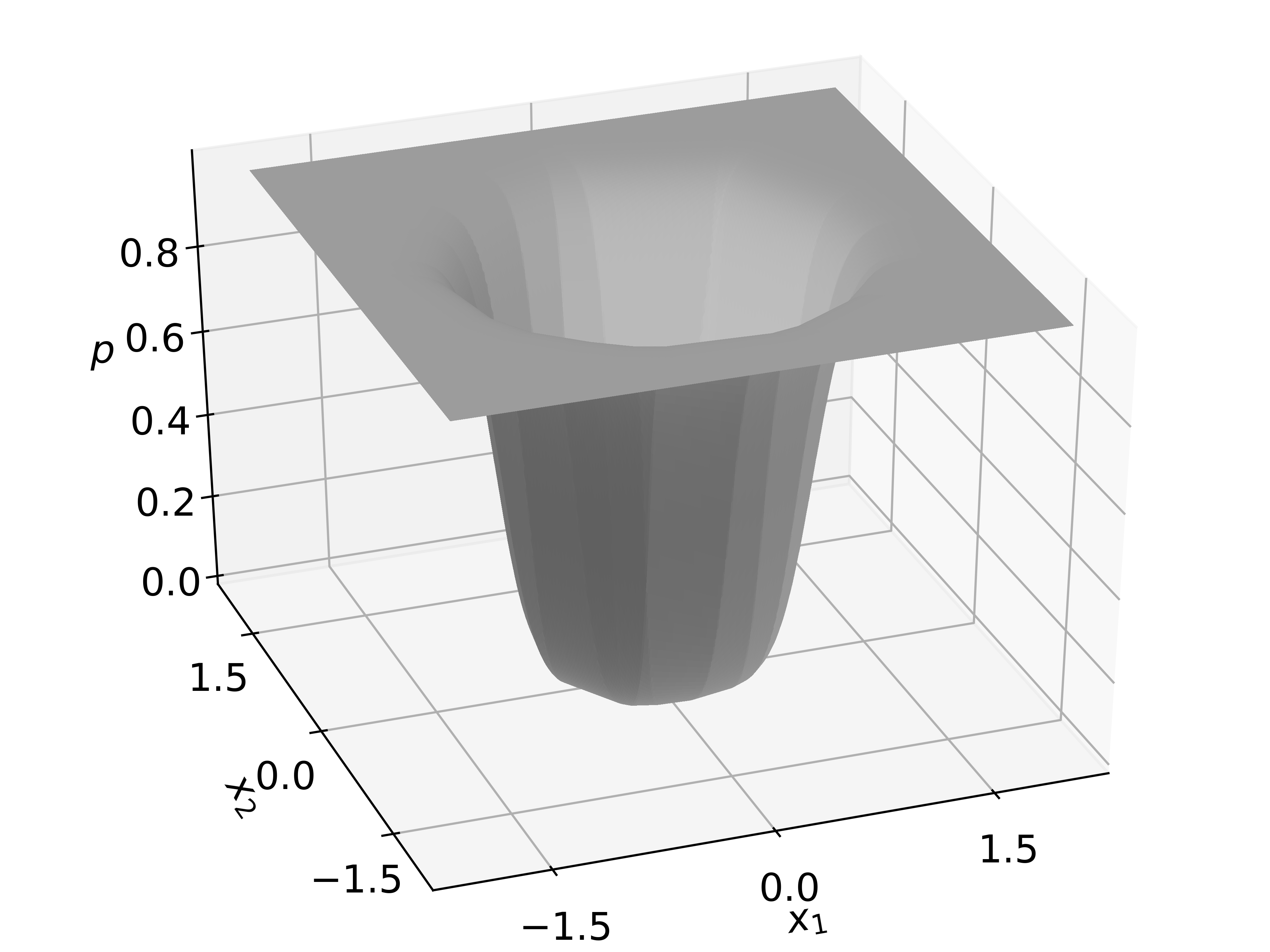} \\
\includegraphics[width=0.45\textwidth]{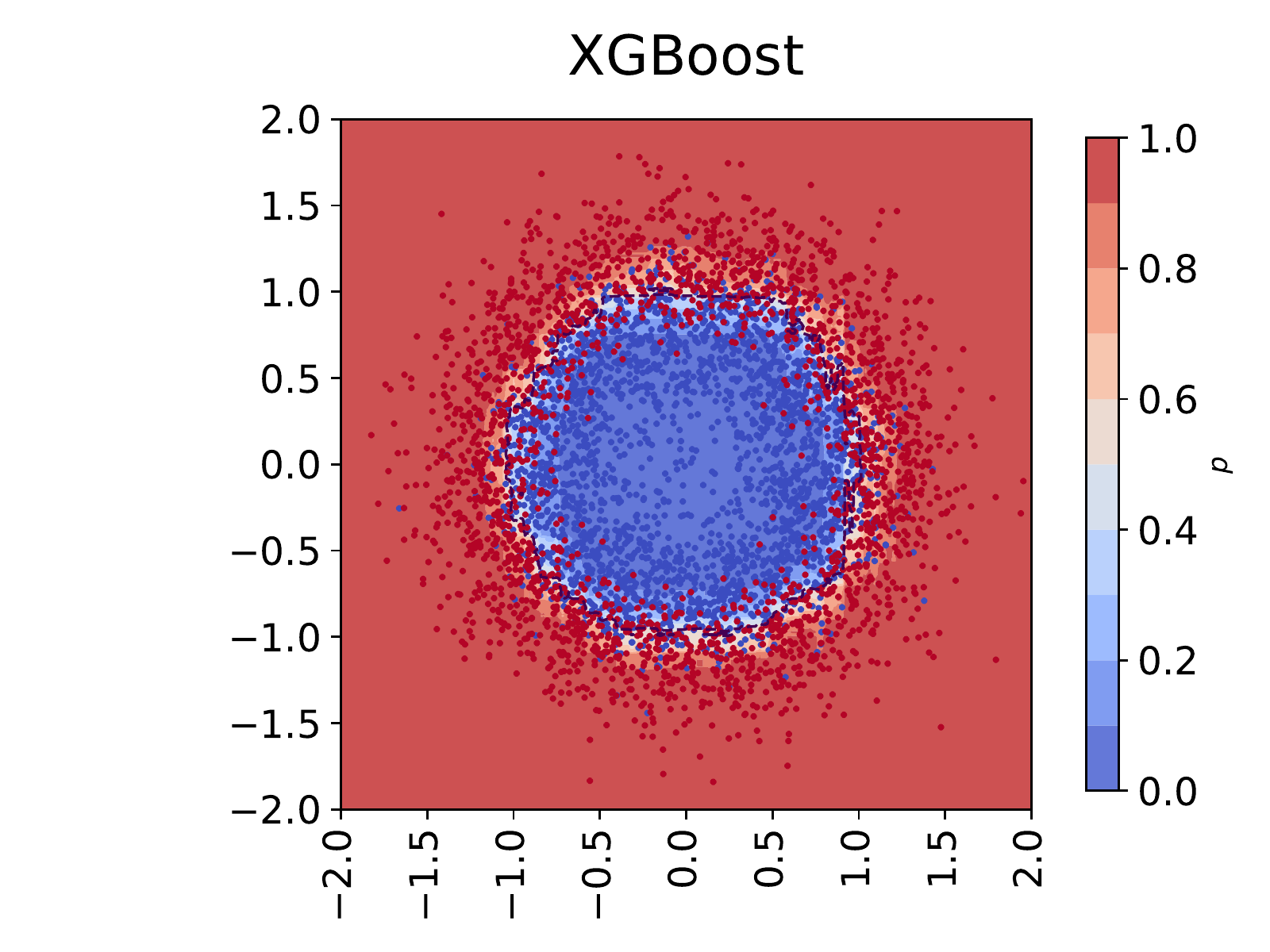}
\includegraphics[width=0.45\textwidth]{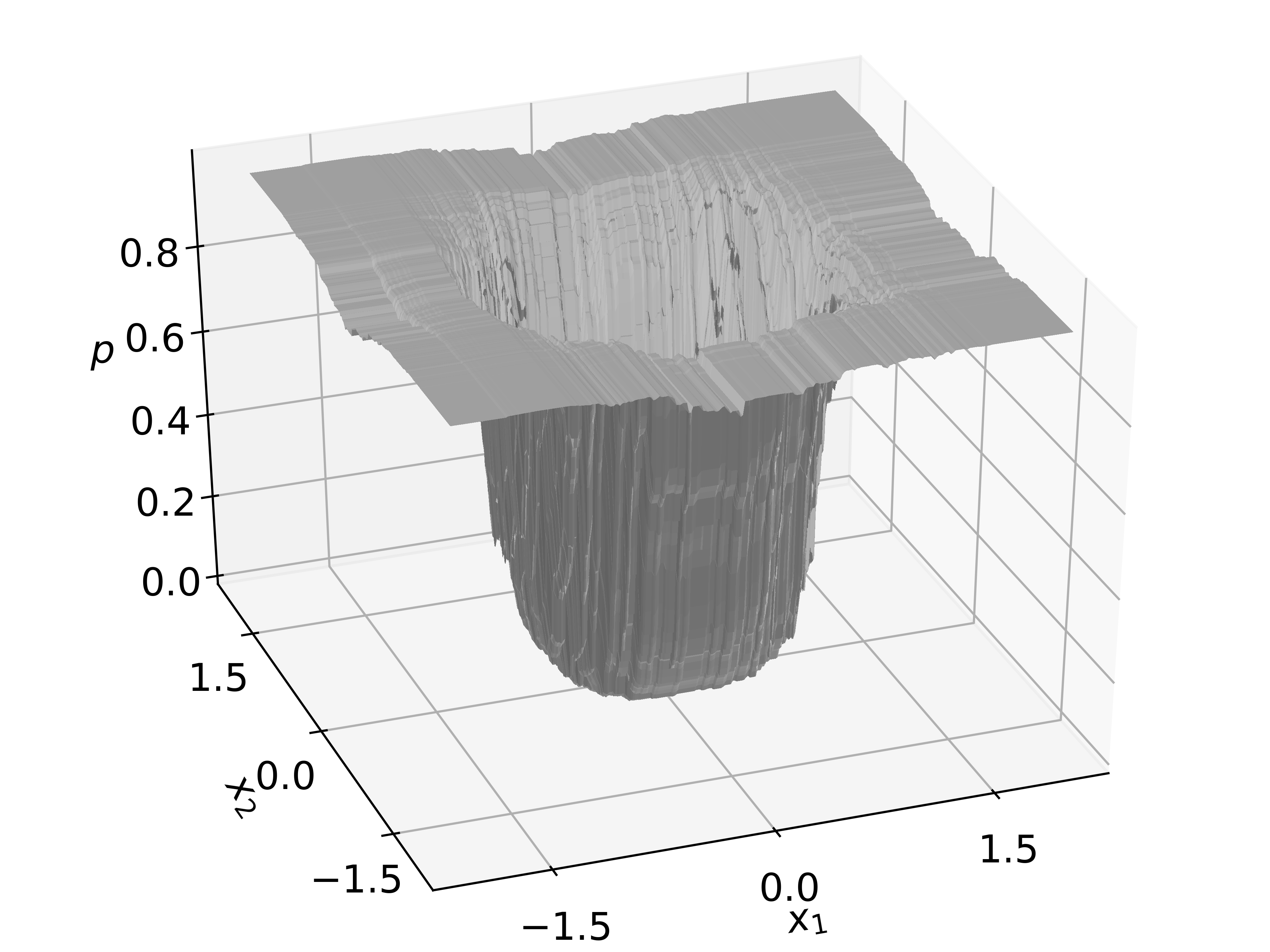}
\caption{Models obtained after training on the sphere data. Blue dots represent class 0 and red dots class 1. The legend numbers indicate the probability of belonging to the red class.} \label{fig:circle}
\end{figure}

In addition, we use the same data to train a decision tree and a neural network using Python's scikit-learn package \cite{SciKit}, and an \xgboost{} classifier. The neural network is fixed to one hidden layer. Figure \ref{fig:circle} shows the resulting models after optimizing the hyperparameters using Optuna, and Table \ref{table:node_results_circle} summarizes the resulting node counts and log-losses. Note that the decision tree, the reticulum and the neural network employ a comparable number of nodes while \xgboost{} requires about two more orders of magnitude.

\begin{table}[!htb]
\centering
\begin{tabular}{lcc}
\toprule
Technique & $|\mathcal{N}|$ & Log-loss \\
\midrule
Decision Tree & 15 & 0.4779 \\
Bayesian Reticulum & 35 & 0.3798 \\
Neural Network & 23 & 0.3511 \\
\xgboost{} & 1\,474 & 0.3630 \\
\midrule
\bottomrule
\end{tabular}
\caption{Number of nodes involving weights $\vw$ and log-loss  for each technique applied to the sphere data set.} \label{table:node_results_circle}
\end{table}

While the decision tree finds a coarse approximation of the sphere that is almost rectangular, the neural network finds the most accurate solution. This is due to the smoothness of the problem which benefits neural networks. The Bayesian Reticulum approximates the circle with slanted planes around the circle boundary, which can be seen as a hybrid between the other two techniques. The \xgboost{} classifier discovers the circle shape but tends to produce a jaggy probability surface. In this case, the neural network outperforms the other methods because of the low noise level.

\afterpage{\clearpage}

\subsection{Recovery of a Cross}
The location of the data for this test is generated uniformly in $[-2,2]\times[-2,2]$, i.e. $\vX=(U_1,U_2)^T$ with two independent distributions $U_1,U_2\sim \text{Uniform}(-2,2)$. Given the locations of $1\,000$ points, the outcomes $Y$ are sampled from a Bernoulli distribution whose probability is 0.9 in the first and third quadrants, and 0.1 otherwise.

\begin{figure}[!h]
\centering
\includegraphics[width=0.45\textwidth]{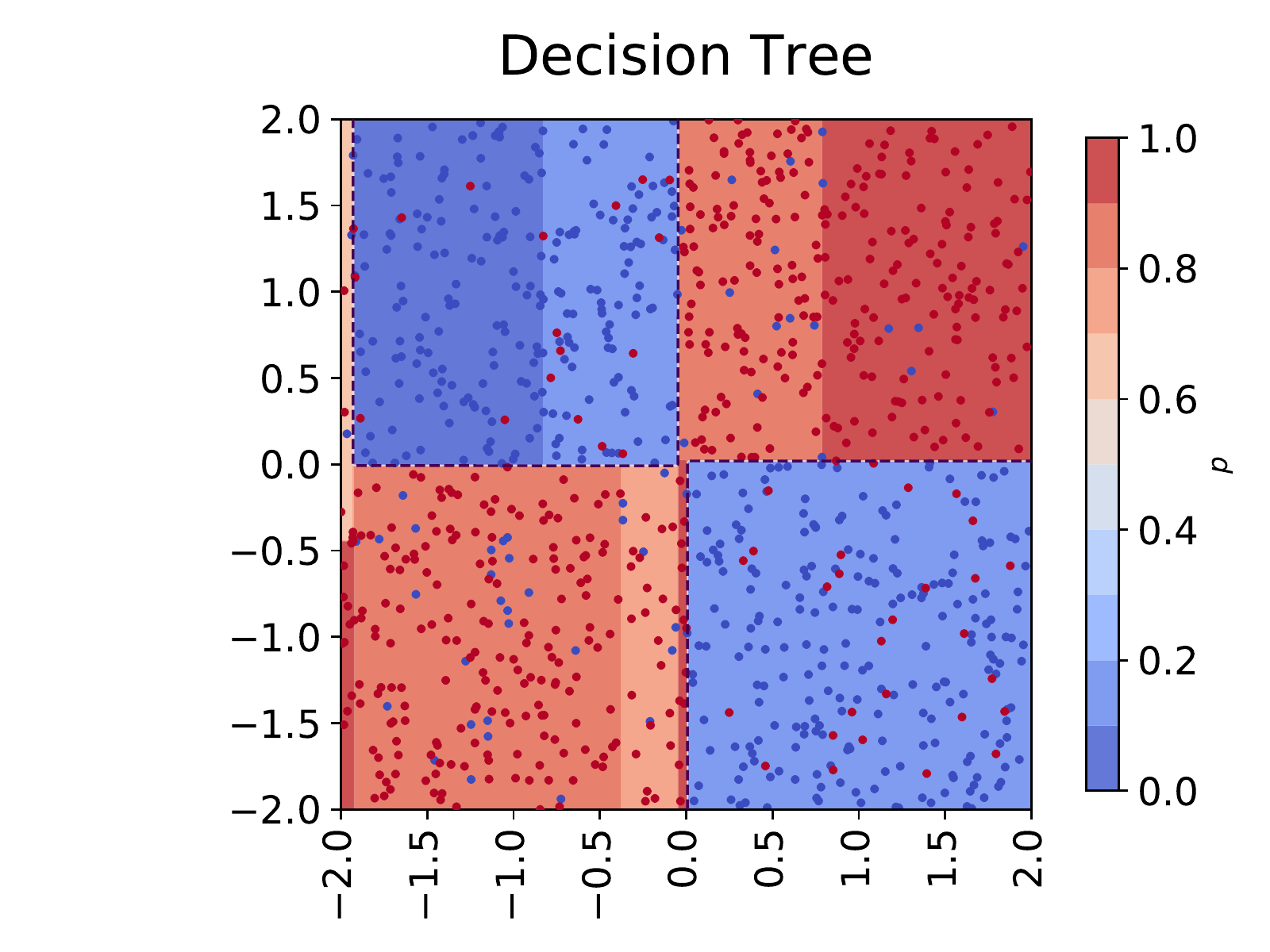}
\includegraphics[width=0.45\textwidth]{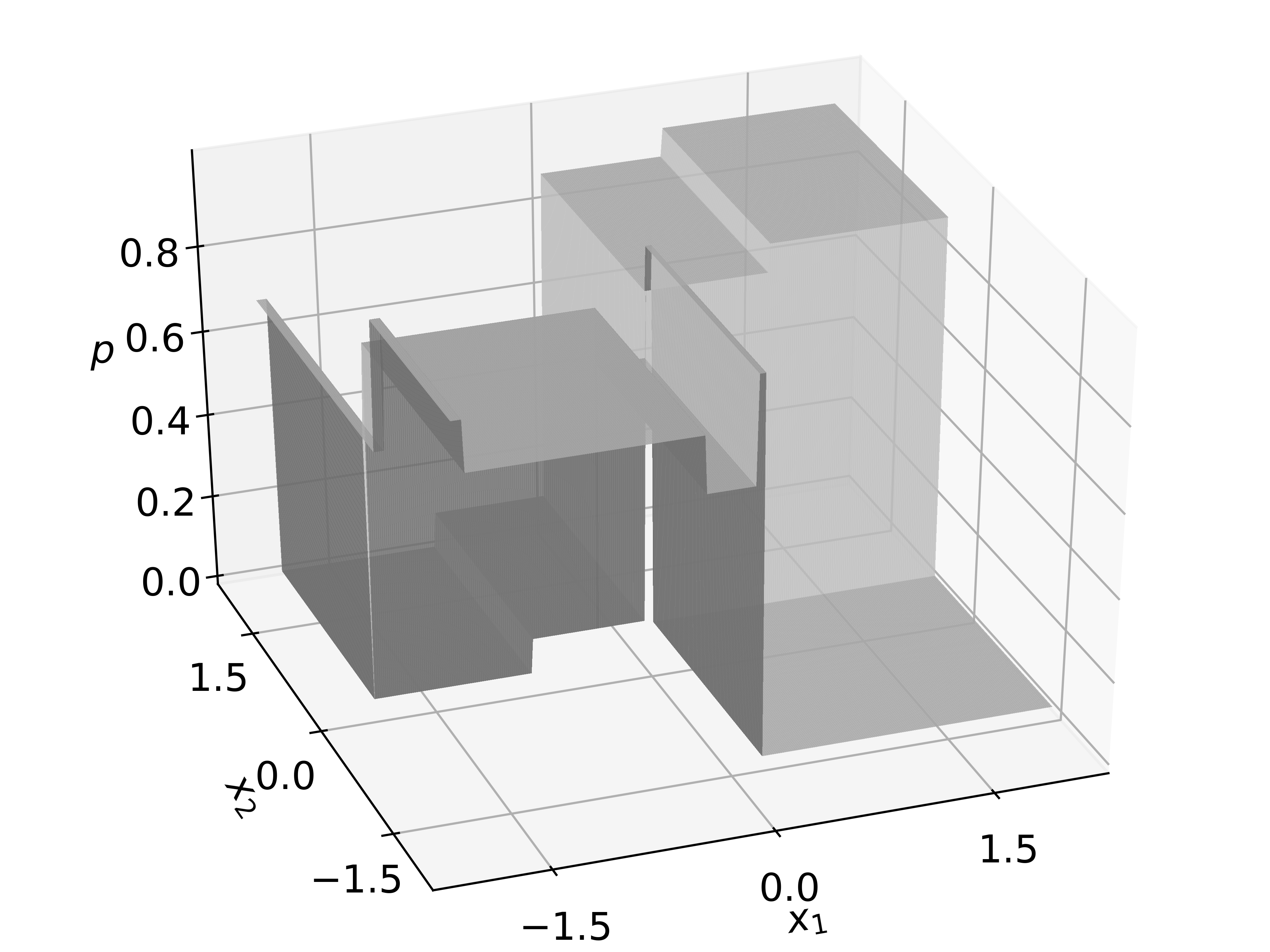} \\
\includegraphics[width=0.45\textwidth]{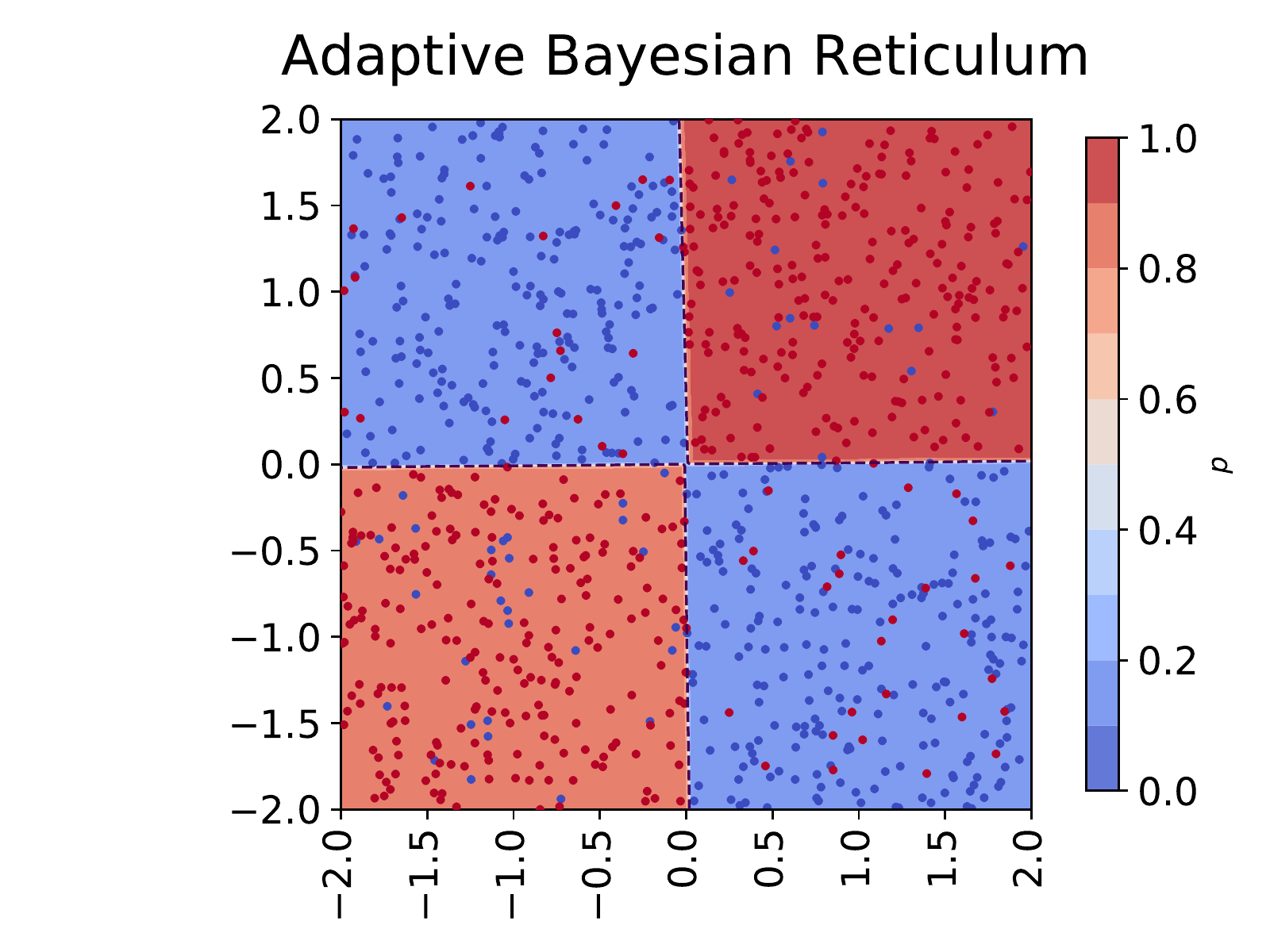}
\includegraphics[width=0.45\textwidth]{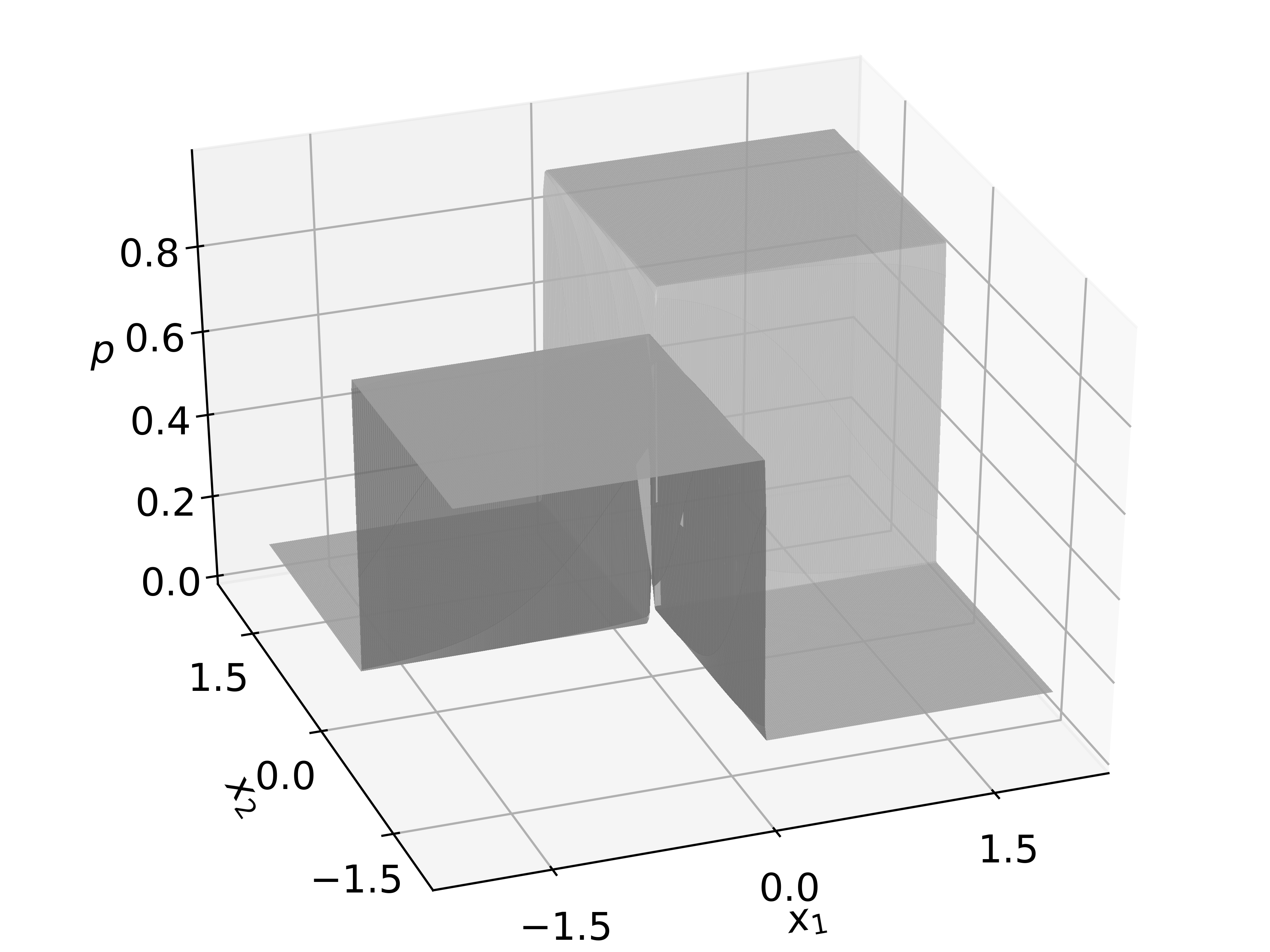} \\
\includegraphics[width=0.45\textwidth]{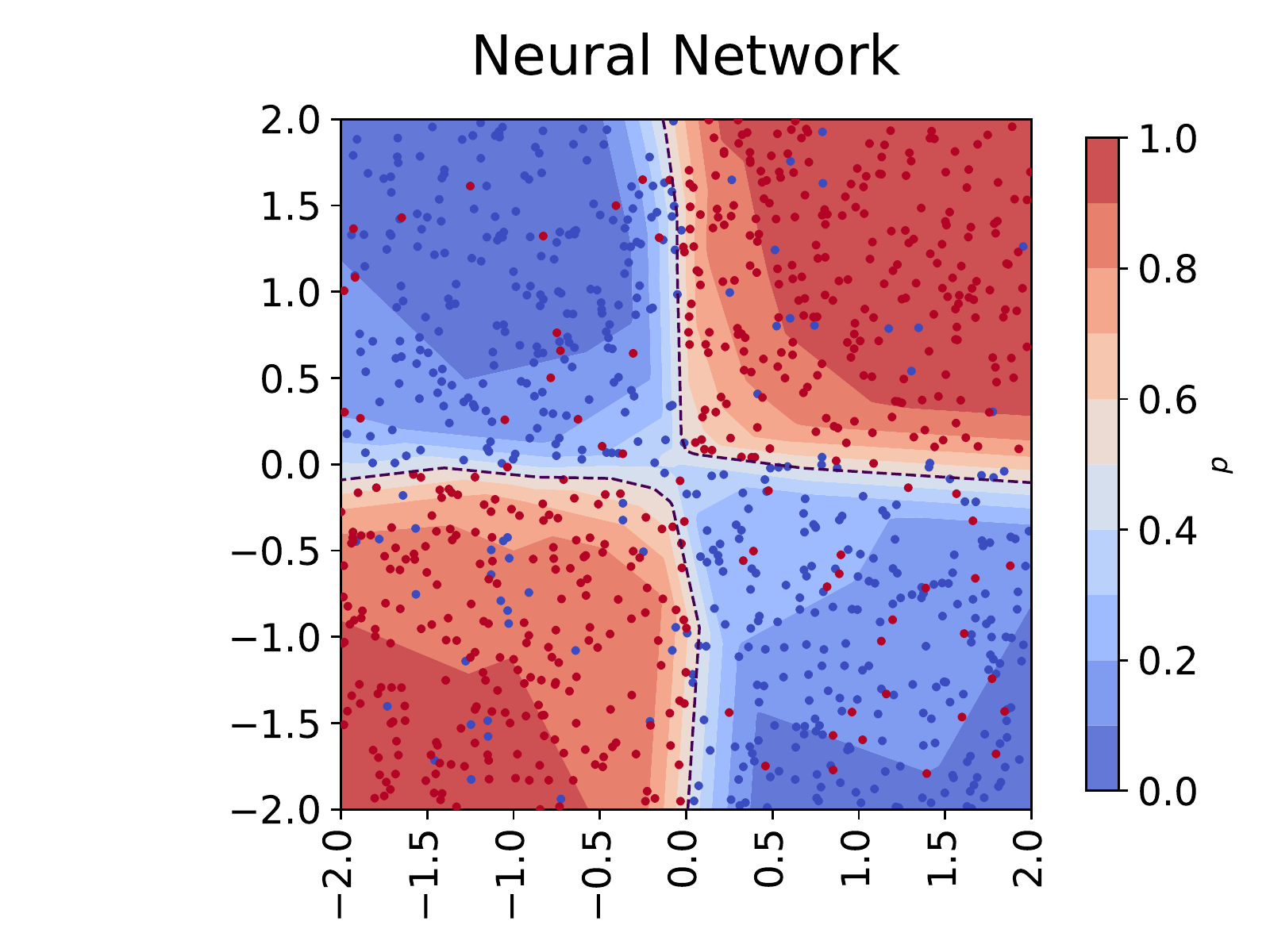}
\includegraphics[width=0.45\textwidth]{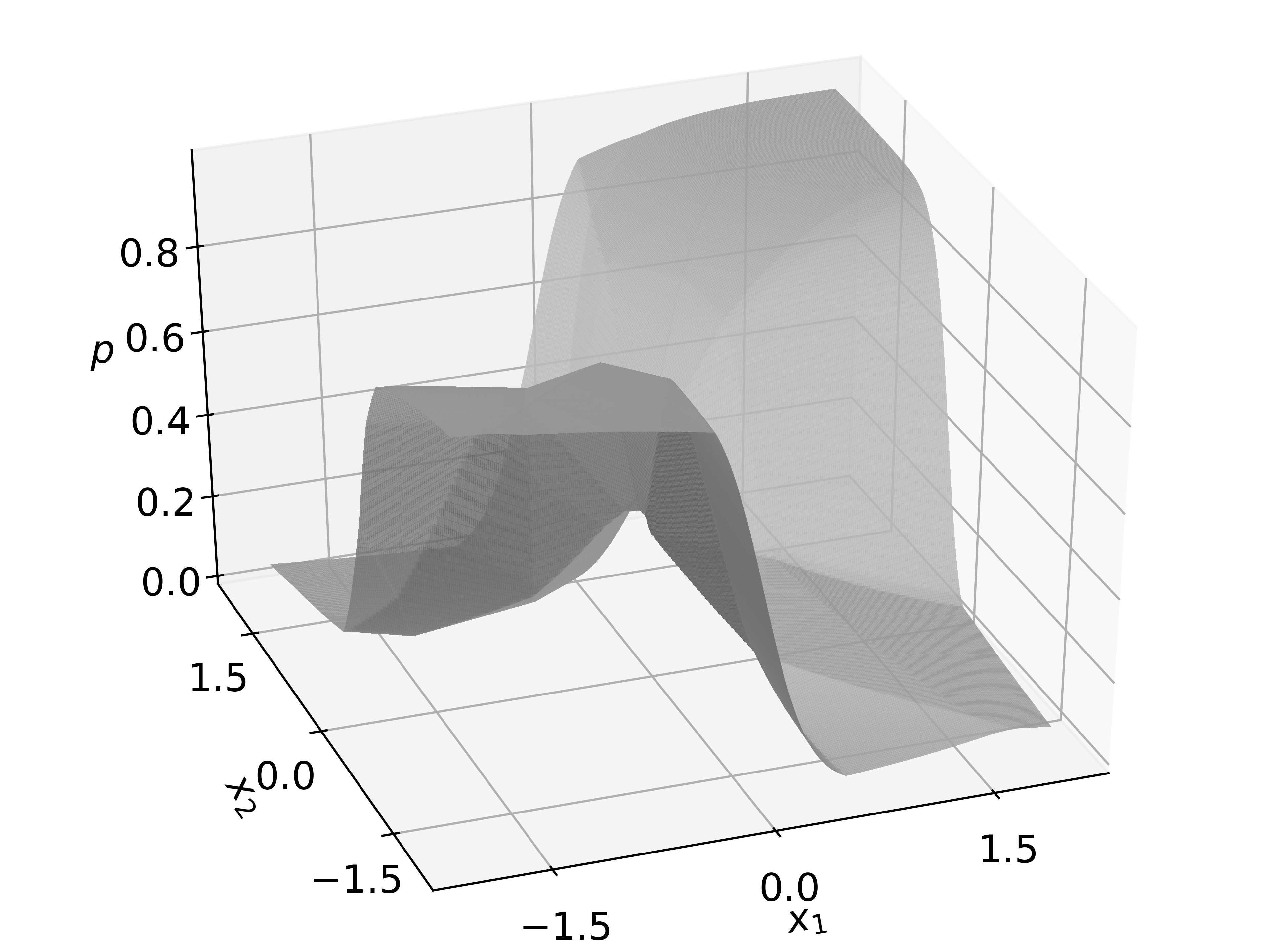} \\
\includegraphics[width=0.45\textwidth]{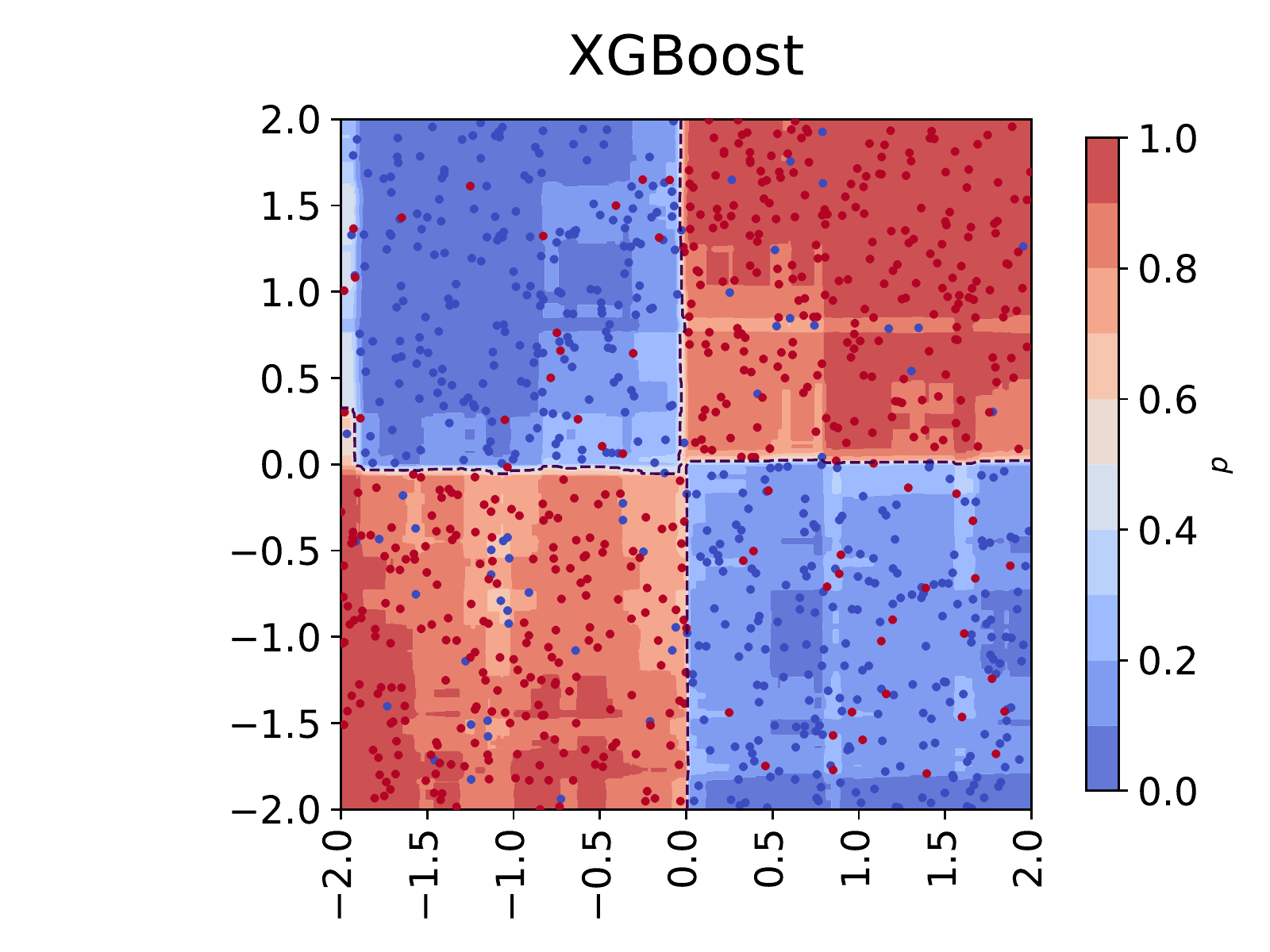}
\includegraphics[width=0.45\textwidth]{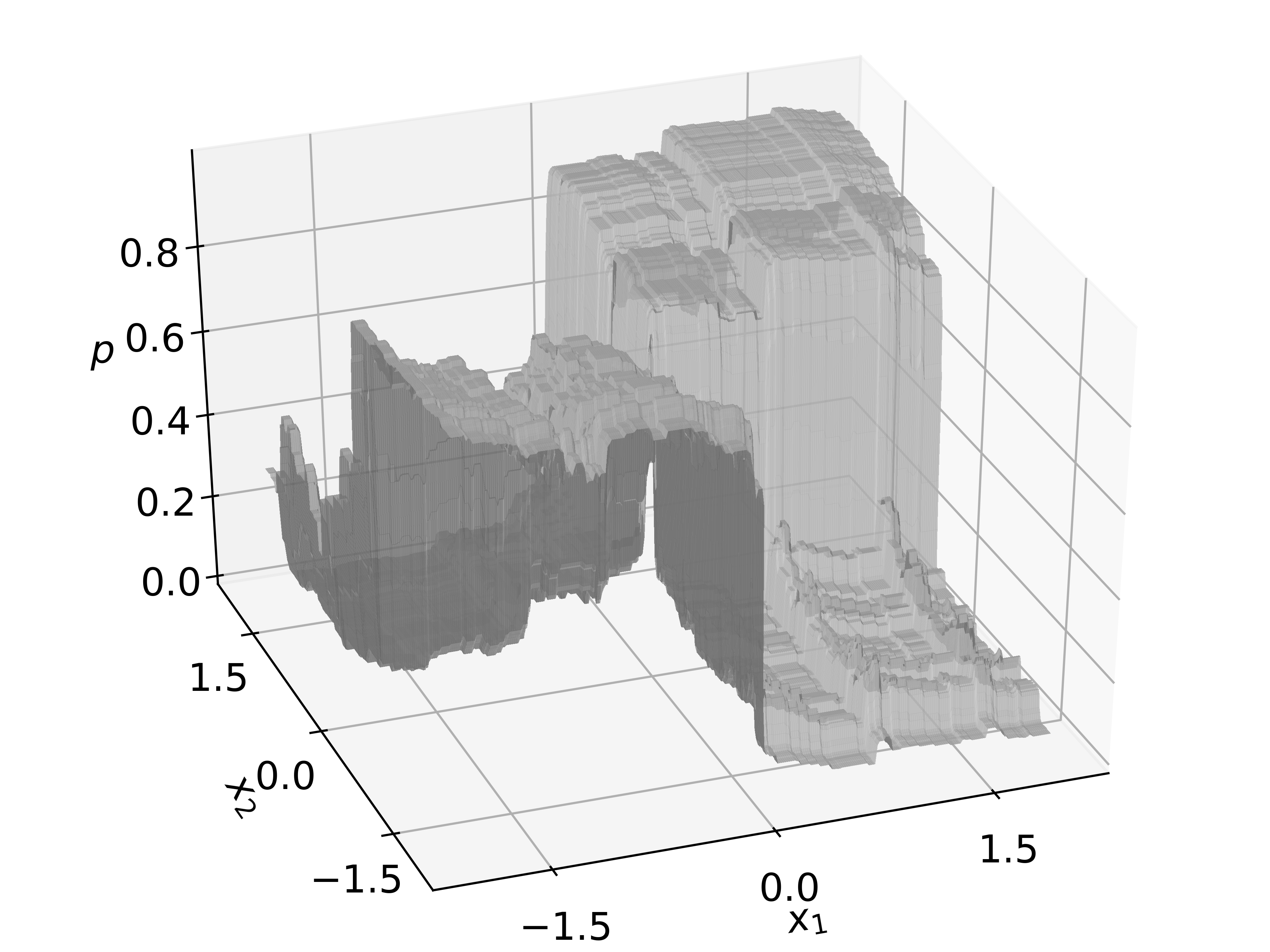}
\caption{Models obtained after training on the cross data. Blue and red points represent classes 0 and 1 respectively.} \label{fig:cross}
\end{figure}

As in the previous example, we compare the trained Bayesian reticulum to a decision tree, a neural network with one hidden layer, and an \xgboost{} classifier. Table \ref{table:node_results_cross} shows the resulting node counts and optimization metrics and Figure \ref{fig:cross} displays the models. Note that the Bayesian reticulum extended to an optimally compact tree with 3 internal nodes, a depth of 2 and 4 leaves representing the quadrants. Both the decision tree and the neural network employ more nodes than the reticulum, leading to some noise overfitting. Finally, \xgboost{} requires a node count more than two orders of magnitude greater than the reticulum.

\begin{table}[!htb]
\centering
\begin{tabular}{lcc}
\toprule
Technique & $|\mathcal{N}|$ & Log-loss \\
\midrule
Decision Tree & 9 & 0.4952 \\
Bayesian Reticulum & 3 & 0.3369 \\
Neural Network & 26 & 0.3677 \\
\xgboost{} & 638 & 0.3345 \\
\midrule
\bottomrule
\end{tabular}
\caption{Number of nodes involving weights $\vw$ and log-loss  for each technique applied to the cross data set.} \label{table:node_results_cross}
\end{table}

In this noisy example, the decision tree, the neural network and \xgboost{} produce unnecessarily complex solutions, i.e. they overfit and find additional patterns on top of the true data generating process. The Bayesian nature of the reticulum is better suited to recover the model for noisy data and produces a compact result. If we only look at the log-loss metric, the optimal technique is \xgboost{} by a small margin. Nonetheless, we know  by construction that \xgboost{} overfits the data and uses a large amount of unnecessary nodes.

\afterpage{\clearpage}

\subsection{Ripley Data Set} \label{sec:ripley}
The Ripley data set, \cite{Rip94a}, is a 2-dimensional binary classification data set generated using a mixture of two Gaussian distributions per class. This data set provides a train and a test set but we ignore the test set in this study. We employ the same 5-fold cross-validation optimization scheme as before and report the best fit.

\begin{figure}
\centering
\begin{subfigure}[c]{0.8\textwidth}
\includegraphics[trim=0 30 0 -20, width=\textwidth]{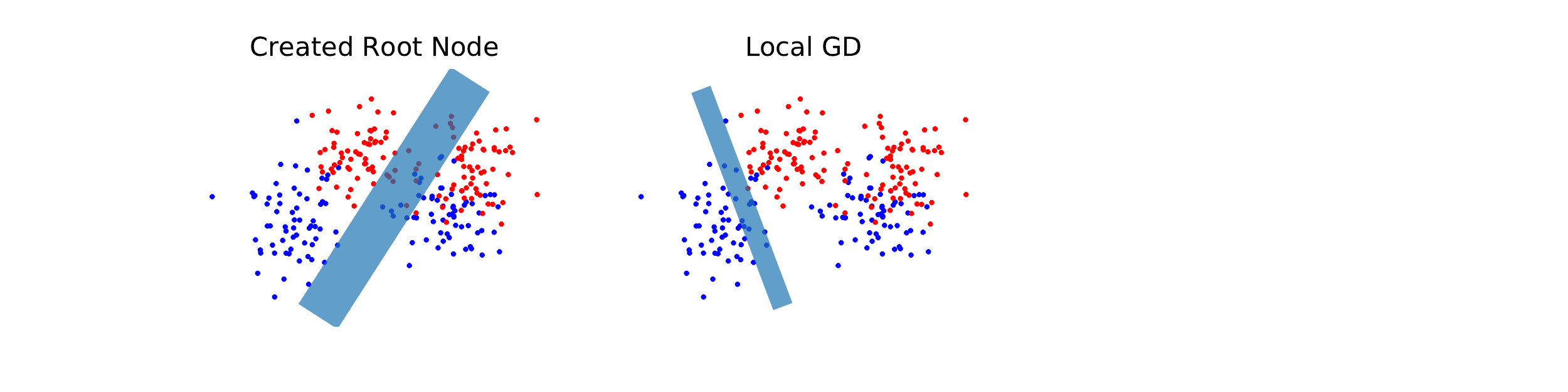}
\subcaption{Initial, random root node split (left) and after local gradient ascent (right).}
\end{subfigure}

\begin{subfigure}[c]{0.8\textwidth}
\includegraphics[trim=0 30 0 -20, width=\textwidth]{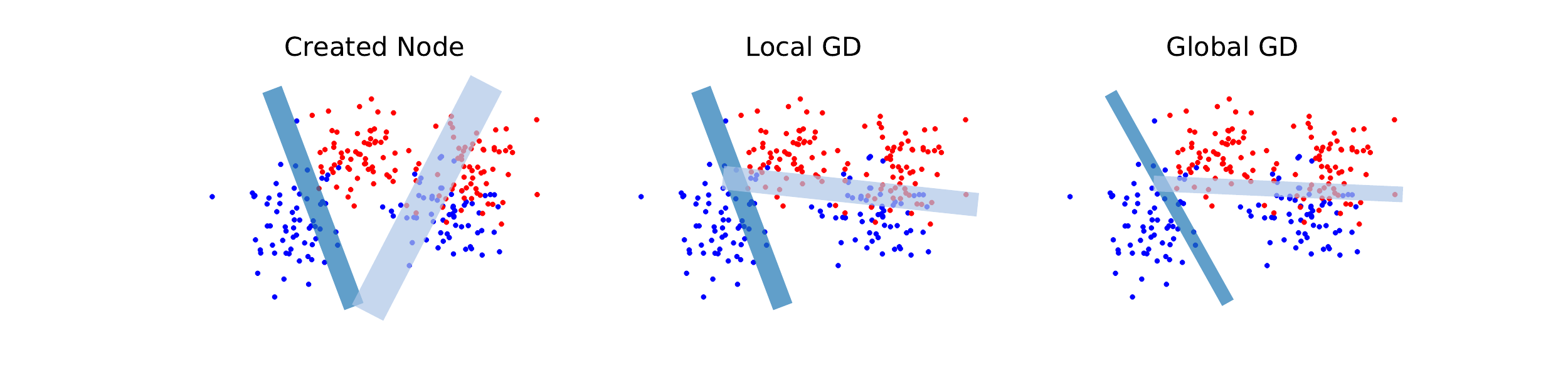}
\subcaption{First child node of the root node just after random initialization (left), after local gradient ascent (middle) and after global gradient ascent (right).}
\end{subfigure}

\begin{subfigure}[c]{0.8\textwidth}
\includegraphics[trim=0 30 0 -20, width=\textwidth]{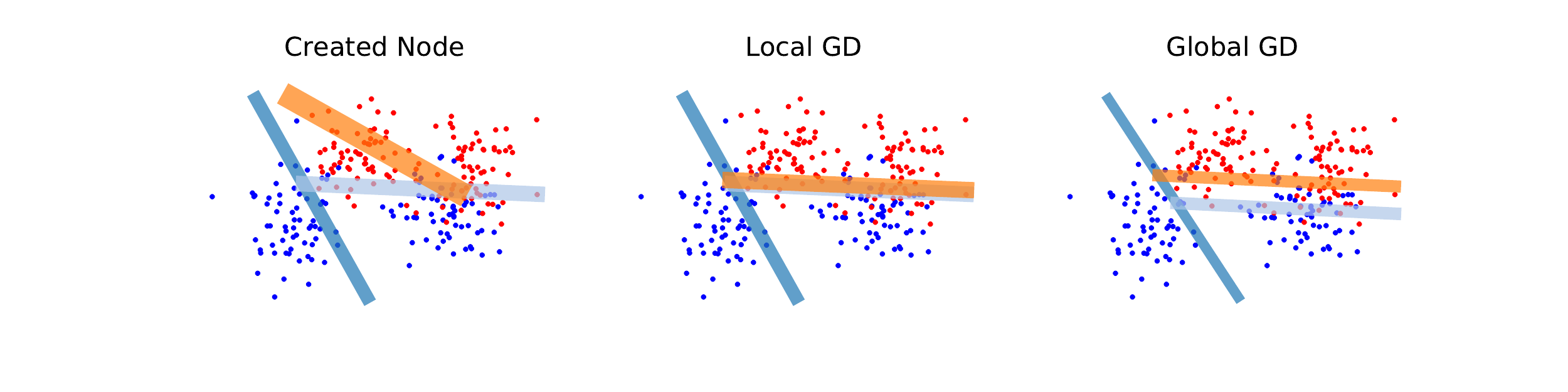}
\subcaption{A child node of the light blue split above at the same three lifecycle stages.}
\end{subfigure}

\begin{subfigure}[c]{0.8\textwidth}
\includegraphics[trim=0 30 0 -20, width=\textwidth]{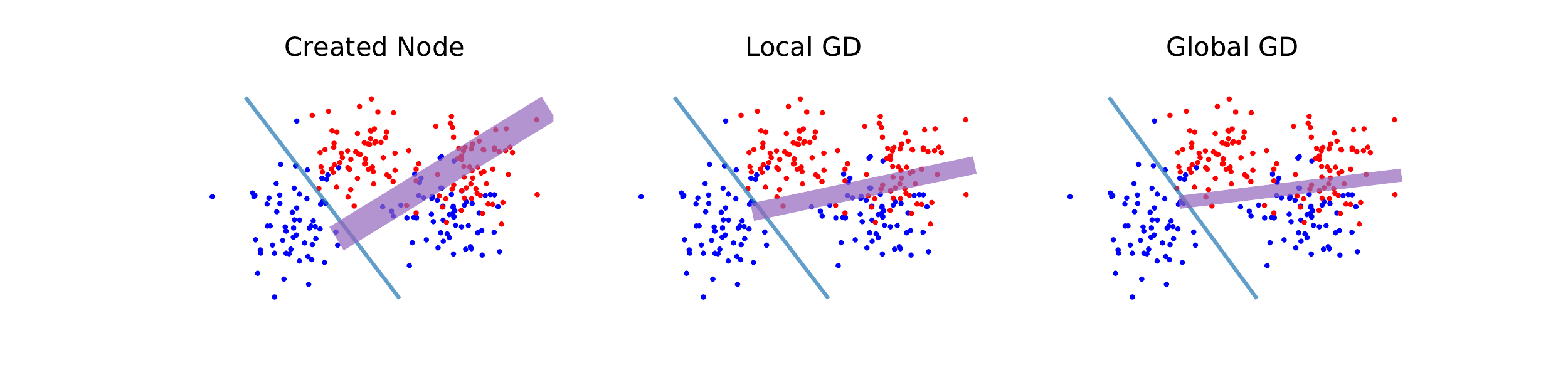}
\subcaption{A few steps further in the tree construction process where the light blue and orange splits have been pruned.}
\end{subfigure}

\begin{subfigure}[c]{0.8\textwidth}
\includegraphics[trim=0 30 0 -20, width=\textwidth]{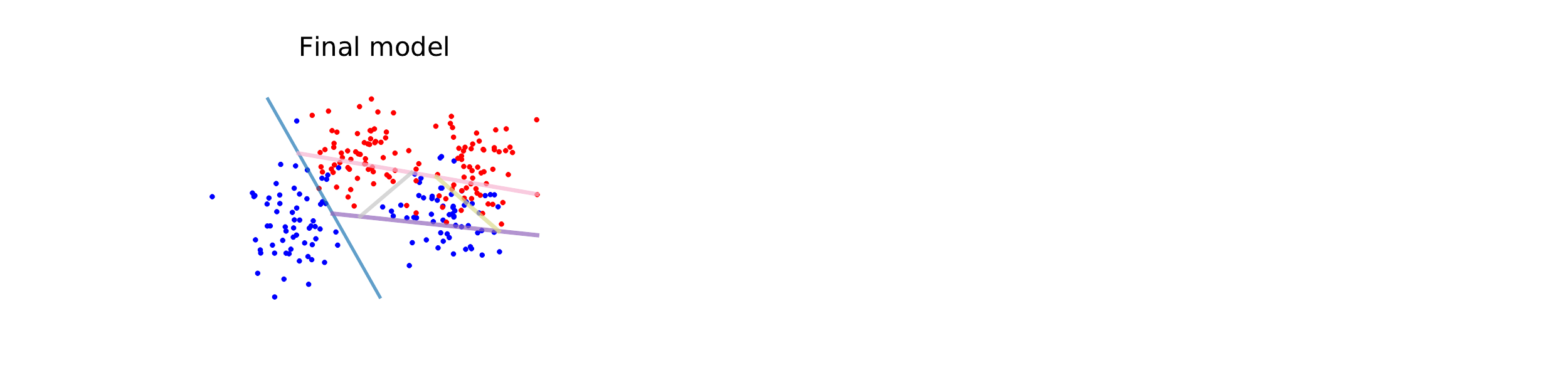}
\subcaption{The final state of the model after completing the training.}
\end{subfigure}

\caption{Construction of a Bayesian Reticulum for the Ripley data set.}
\label{fig:ripley_construction}
\end{figure}

We use this data set to give an insight into the Bayesian Reticulum model construction process at various stages during training, see Figure \ref{fig:ripley_construction}. Each row from (a) to (d) shows the evolution of a newly created node from its random initial position (left) to its state after local gradient ascent (middle; after only the newly added node has been optimized) and finally to the position after global gradient ascent (right; after all node weights have been optimized). Row (a) shows the root node construction, row (b) the root's first child and row (c) the child of the previously added child. Row (d) shows the model construction process a few steps later in the process where both the light blue and the orange splits have been pruned and a new, purple split has been added. Row (e) shows the final state of the model after training completion. Wide split lines correspond to a wide transition period of the sigmoidal split corresponding to softness, whereas thin lines represent stiff splits closer to a Heaviside step function. One can clearly see that soft splits adjust their orientation quickly and significantly during the early gradient ascent stages. With time they become stiffer until the end of model training.

\begin{figure}
\centering
\includegraphics[width=0.45\textwidth]{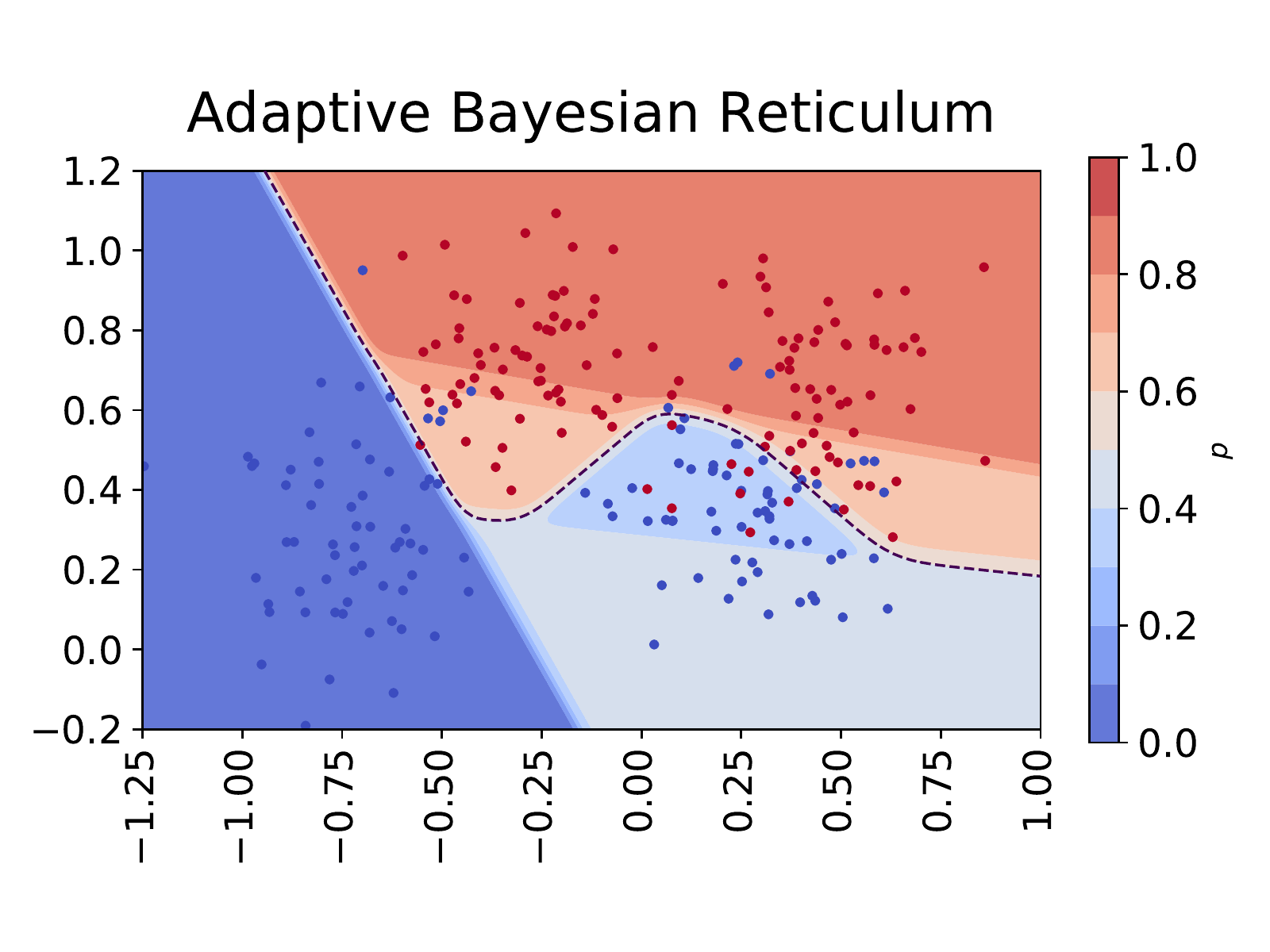}
\includegraphics[width=0.45\textwidth]{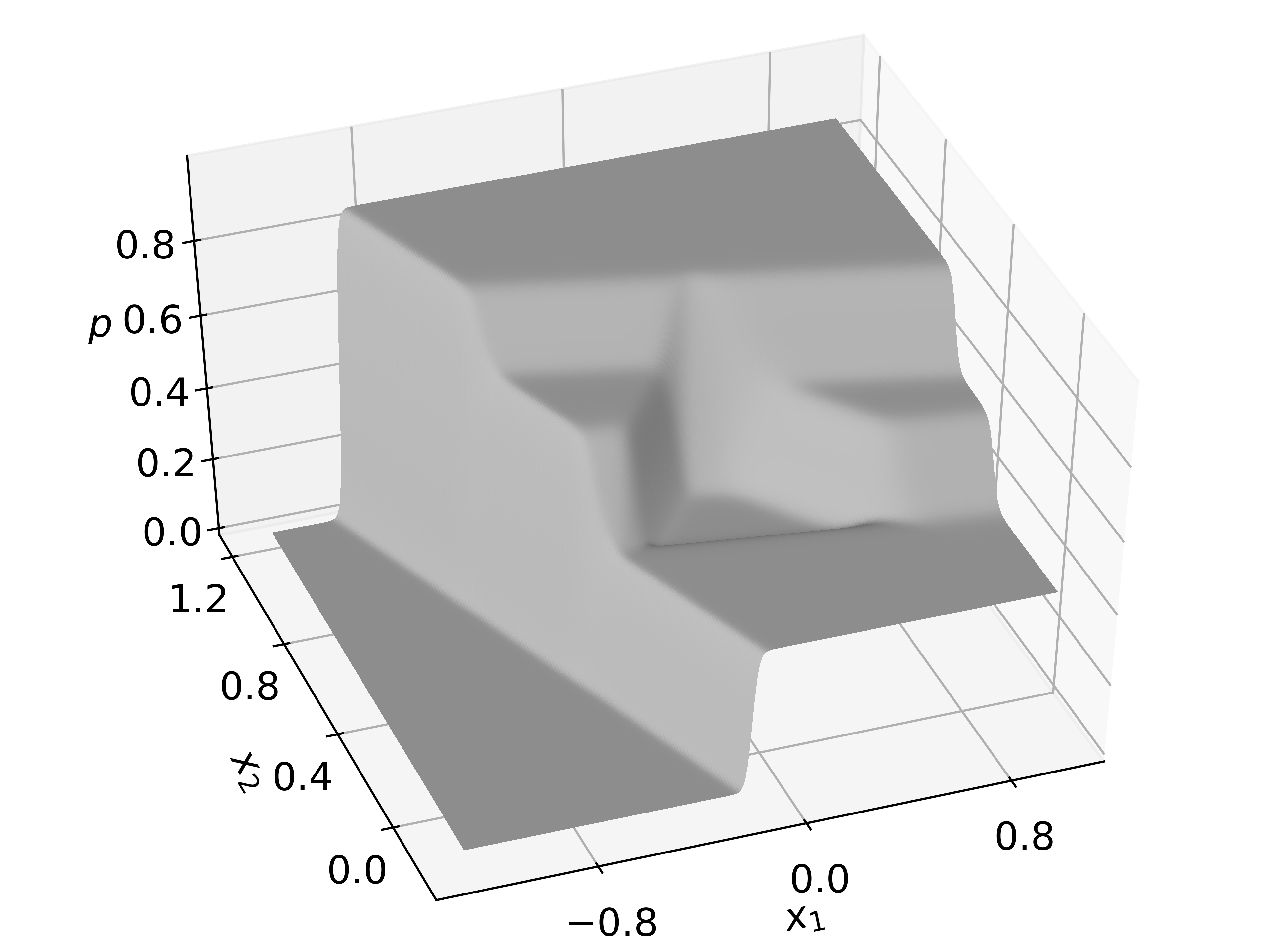}
\caption{The Bayesian Reticulum resulting from training on the Ripley data set. The blue and red points represent classes 0 and 1 respectively.}
\label{fig:rip_ret}
\end{figure}

Figure \ref{fig:rip_ret} visualizes the probability surface of the resulting model. Note that the splits correspond to the final splits in Figure \ref{fig:ripley_construction}.

\section{Conclusive Remarks}

Bridging decision trees and neural networks exposes a number of interesting aspects of both techniques alongside a novel classification technique that has the advantages of both: a soft tree structure that can be extended (and pruned) iteratively with slanted hyperplanes optimized via gradient ascent, first locally and then globally.\\

One of the key concepts is the idea of a node's unexplained potential to guide the search of the optimal topology, moving us closer to being able to model a simple problem with a simple topology (following Occam's Razor). In our opinion this will likely be one of the most interesting areas of research, where the unexplained potential is but one of the important aspects of automatically building a network: which two nodes to connect (in the standard neural network formulation), and how to model cyclical graphs. Furthermore, we show how the use of the beta function to regularize the model gives a natural stopping condition to the search for a more complex topology which accounts for the available statistical evidence (albeit via an approximation).\\

Finally, interpreting the weights in polar coordinates allows for a more explainable model, at least in the tree formulation with a gate function. In future research we would like to express the problem in polar coordinates which enables us to directly model the angle(s) of the hyperplane, its intercept and the stiffness of the sigmoid curve --- opening an interesting area of research for explainable neural networks.

\newpage
\appendix
\section{Implementation Details} \label{app:implementation}
This section explains in more detail some of the steps of algorithm \ref{alg:reticulum_algo}:
\begin{itemize}
\item Step \ref{algo:line:sample_weights}, initial weights sampling: In order to properly initialize the weights we first need to establish an outlier-resistant data range estimate such that we can appropriately scale the weight vector to reflect the desired initial stiffness, which is a model parameter, across the data set. This is being done by computing the difference between the 0.25 and 0.75 quantiles in each dimension, which we call pseudo-range. Note that only the root node will use all the data for this process; subsequent child nodes will only compute the pseudo-range on data points where the incoming probability from the parent node is $>0.5$, simulating a hard split higher up the hierarchy. Then, we sample a uniformly distributed point from the unit hypersphere corresponding to the problem dimension, which we interpret as a weight vector of length 1. We then divide the vector's entries by the pseudo-range for each feature dimension, which is equivalent to scaling the data to a fixed range. Finally, we center the vector at the median of the data points.
\item Steps \ref{algo:line:train1} and \ref{algo:line:train2}, gradient ascent: We employ the standard Adam optimizer of \cite{Adam14} algorithm which implements automatic step-size adaption for each problem dimension. The initial step size and the number of gradient ascent steps are model parameters. Half of the latter will be used in the local optimization and half in the global one.
\item Step \ref{algo:line:prune}, pruning: This step requires a model parameter called \emph{pruning factor}. A node is being pruned of its children whenever the sum of the children values $c_\ell$ is not better than its parent $c_\ell$ plus $\ln((\text{pruning factor})^{\text{level} + 1})$, where the nesting level starts with 0 at the root level and the pruning factor should be a value in $[1, 1.2]$.
\end{itemize}

\section{Backpropagation} \label{app:backprop}
Our goal is to maximize equation \eqref{eq:approx_loglike}. We provide its gradient in order to apply a gradient ascent algorithm. The derivative of $c_\ell$ with respect to the weight $w_{k,j'}$  is,
\[
\frac{\partial c_\ell}{\partial w_{k,j'}} = \sum_{z, i, j} \frac{\partial \ln(B)}{\partial z} \frac{\partial z}{\partial s_j^i} \frac{\partial s_j^i}{\partial w_{k,j'}}.
\]
In the summation above, $z\in\{\alpha',\beta'\}$, $i=1,\dots,m$, and $j$ is the node index. We use the notation $w_{k,j'}$ to indicate the weight for dimension $k$ and node index $j'$. Similarly, we will also use $x_{k,i}$ to indicate the coordinate value of dimension $k$ for point $\vx_i$.

The derivative of $\ln(B)$,
\[
\frac{\partial \ln(B)}{\partial z} = \psi(z) - \psi(\alpha' + \beta'),
\]
where $\psi$ is the digamma function. Then,
\[
\frac{\partial z}{\partial s_j^i} = 
\begin{cases}
\frac{\partial p(\vx_i \in \ell)}{\partial s_j^i}(1 - y_i) & \text{ if } z = \alpha', \\
\frac{\partial p(\vx_i \in \ell)}{\partial s_j^i}y_i & \text{ if } z = \beta'.
\end{cases}
\]
The derivative $\partial p(\vx_i \in \ell) / \partial s_j^i$ is 0 if $j$ is not the index of a predecessor of $\ell$. Otherwise, it is $p(\vx_i \in \ell) / s_j^i$ if $j$ is the index of a left child, or $-p(\vx_i \in \ell) / s_j^i$ if $j$ is the index of a right child. Note that $p(\vx_i \in \ell) / s_j^i$ can be computed as $p(\vx_i \in \ell)$ omitting the weight corresponding to $s_j^i$. According to the mapping described in Appendix \ref{app:implementation}, to verify whether $j$ corresponds to a left or a right child, one can simply verify whether $j$ is an odd number. Finally,
\begin{equation} \label{eq:partial_a_w}
\frac{\partial s_j^i}{\partial w_{k,j'}} = 
\begin{cases}
s_j^i (1 - s_j^i)x_{k,i} & \text{ if } j = j', \\
0 & \text{ otherwise}.
\end{cases}
\end{equation}

We can perform a backpropagation in three steps: 1) we start at the root, i.e. $j=0$, and move forward to compute $p(\vx_i \in \ell) / s^i_j$ for each node; 2) once a leaf is reached, we can compute and return $(1-y_i)\partial \ln(B)/\partial \alpha' + y_i\partial \ln(B) / \partial \beta' $; 3) we move backwards, and compute the sum of the left and right children weighted by $p(\vx_i \in \ell) / s^i_j$, and $-p(\vx_i \in \ell) / s^i_j$ respectively. This sum is used to complete the partial derivatives with respect to the weights at $j$. We return the sum of the children weighted by $s^i_j$ and $1 - s^i_j$ respectively to account for the weights $s^i_j$ in the derivatives of the predecessors of $j$.

\section{Polar Coordinates In Neural Networks} \label{app:polar}
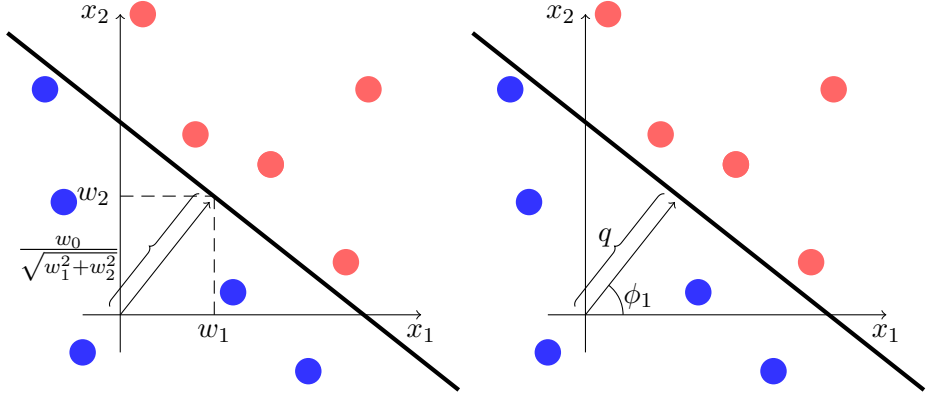
\begin{figure}
\centering
\begin{tikzpicture}
    \coordinate (O) at (0,0) ;
    \coordinate (A) at (0,4) ;
    \coordinate (B) at (0,-0.5) ;
    \coordinate (C) at (4,0) ;
    \coordinate (D) at (-0.5,0) ;
    \coordinate (plane) at (2.375/2, 3/2);

    \node[left] at (A) {$x_2$};
    \draw[->] (B) -- (A) ;
    \node[below] at (C) {$x_1$};
    \draw[->] (D) -- (C) ;

    \node[circle,fill=red!60,inner sep=0pt,minimum size=10pt] at (2,2) {};
    \node[circle,fill=red!60,inner sep=0pt,minimum size=10pt] at (1,2.4) {};
    \node[circle,fill=red!60,inner sep=0pt,minimum size=10pt] at (0.3,4) {};
    \node[circle,fill=red!60,inner sep=0pt,minimum size=10pt] at (3,0.7) {};
    \node[circle,fill=red!60,inner sep=0pt,minimum size=10pt] at (3.3,3) {};
    \node[circle,fill=red!60,inner sep=0pt,minimum size=10pt] at (2,2) {};

    \node[circle,fill=blue!80,inner sep=0pt,minimum size=10pt] at (-0.5,-0.5) {};
    \node[circle,fill=blue!80,inner sep=0pt,minimum size=10pt] at (1.5,0.3) {};
    \node[circle,fill=blue!80,inner sep=0pt,minimum size=10pt] at (-1,3) {};
    \node[circle,fill=blue!80,inner sep=0pt,minimum size=10pt] at (-0.75,1.5) {};
    \node[circle,fill=blue!80,inner sep=0pt,minimum size=10pt] at (2.5,-0.75) {};

    \draw[->] (O) -- (plane);
    \draw[,ultra thick] (4.5,-1) -- (-1.5,3.75);
    \draw[dash pattern=on5pt off3pt] (2.375/1.9, 0) -- (2.375/1.9, 3/1.9);
    \draw[dash pattern=on5pt off3pt] (0, 3/1.9) -- (2.375/1.9, 3/1.9);
    \node[below] at (2.375/1.9, 0)  {$w_{1}$};
    \node[left] at (0, 3/1.9)  {$w_{2}$};

    \draw[decoration={brace,mirror,raise=5pt},decorate]
		  (plane) -- node[above=0pt,left=13pt] {$\frac{w_0}{\sqrt{w_1^2+w_2^2}}$} (0,0);
\end{tikzpicture}
\begin{tikzpicture}
    \coordinate (O) at (0,0) ;
    \coordinate (A) at (0,4) ;
    \coordinate (B) at (0,-0.5) ;
    \coordinate (C) at (4,0) ;
    \coordinate (D) at (-0.5,0) ;
    \coordinate (plane) at (2.375/2, 3/2);
   
    \node[left] at (A) {$x_2$};
    \draw[->] (B) -- (A) ;
    \node[below] at (C) {$x_1$};
    \draw[->] (D) -- (C) ;

    \node[circle,fill=red!60,inner sep=0pt,minimum size=10pt] at (2,2) {};
    \node[circle,fill=red!60,inner sep=0pt,minimum size=10pt] at (1,2.4) {};
    \node[circle,fill=red!60,inner sep=0pt,minimum size=10pt] at (0.3,4) {};
    \node[circle,fill=red!60,inner sep=0pt,minimum size=10pt] at (3,0.7) {};
    \node[circle,fill=red!60,inner sep=0pt,minimum size=10pt] at (3.3,3) {};
    \node[circle,fill=red!60,inner sep=0pt,minimum size=10pt] at (2,2) {};

    \node[circle,fill=blue!80,inner sep=0pt,minimum size=10pt] at (-0.5,-0.5) {};
    \node[circle,fill=blue!80,inner sep=0pt,minimum size=10pt] at (1.5,0.3) {};
    \node[circle,fill=blue!80,inner sep=0pt,minimum size=10pt] at (-1,3) {};
    \node[circle,fill=blue!80,inner sep=0pt,minimum size=10pt] at (-0.75,1.5) {};
    \node[circle,fill=blue!80,inner sep=0pt,minimum size=10pt] at (2.5,-0.75) {};

    \draw[->] (O) -- (plane);
    \draw[,ultra thick] (4.5,-1) -- (-1.5,3.75);

    \draw (0.5,0) arc (0:51.63:0.5);
    \node[] at (20:0.75)  {$\phi_{1}$};
    \draw[decoration={brace,mirror,raise=5pt},decorate]
		  (plane) -- node[above=8pt,left=3pt] {$q$} (0,0);
\end{tikzpicture}
\caption{A hyperplane defined in Cartesian coordinates on the left, and polar coordinates on the right. While in Cartesian coordinates we define the hyperplane as $w_0 + w_{1}x_1 + w_2x_2=0$, in polar we can write $q + \cos(\phi_1)x_1 + \sin(\phi_1)x_2=0$.}\label{fig:polar_example}
\end{figure}

In a bid to make the model explainable, we note that the node weights $\vw$ have a geometrical interpretation. For a hyperplane $w_0 + \sum_{k=1}^d w_kx_k = 0$, the weights $w_1, \dots, w_d$ define the hyperplane normal vector, and $w_0/r$ defines the distance or location of the hyperplane to the origin with $r = (\sum_{k=1}^d w_k^2)^{1/2}$. Figure \ref{fig:polar_example} shows how a hyperplane in $\reals^2$ is defined in Cartesian and polar coordinates.

The Euclidean distance of a point $\vx$ to the hyperplane defined by $\vw$ is $|w_0 + \sum w_kx_k| / r$. This is related to the signed distance introduced in Section \ref{sec:adp_reticulum}, $d_{\calM}({\vx}) = w_0 + \sum w_kx_k$. If we disregard the sign, $d_{\calM}({\vx})$ is the Euclidean distance times $r$. This scaling is not accidental and defines the softness of our gate function margin. Notice that a large scaling will transform the sigmoid $s$ into the Heaviside function.

Given some weights $\vw$, the polar substitution is in particular
\begin{equation} \label{eq:substitution}
\begin{split}
w_0 & = rq, \\
w_1 & = r\cos(\phi_1), \\
w_2 & = r\sin(\phi_1)\cos(\phi_2), \\
\vdots & \\
w_{d-1} & = r\sin(\phi_1)\cdots\sin(\phi_{d-2})\cos(\phi_{d-1}), \\
w_{d} & = r\sin(\phi_1)\cdots\sin(\phi_{d-2})\sin(\phi_{d-1}),
\end{split}
\end{equation}
for $q\in\reals$, $r > 0$, $\phi_1,\dots,\phi_{d-2}\in[0,\pi]$, and $\phi_{d-1}\in[0,2\pi)$. The signed distance from Section \ref{sec:adp_reticulum} becomes
\begin{equation} \label{eq:inner_polar}
d_{\calM}({\vx}) = r[q+\cos(\phi_1)x^1 +\cdots + \sin(\phi_1)\cdots\sin(\phi_{d-1})x^d].
\end{equation}
This substitution factorizes the sigmoid scaling $r$. If we use the polar parametrization, one needs to multiply the gradient matrix from Appendix \ref{app:backprop} on the left by the transposed Jacobian of \eqref{eq:substitution}.

\bibliography{adaptive_reticulum}
\bibliographystyle{ieeetr}

\end{document}